\documentclass[11pt]{article}

\usepackage[preprint]{acl}

\usepackage{times}
\usepackage{latexsym}

\usepackage[T1]{fontenc}

\usepackage[utf8]{inputenc}

\usepackage{microtype}

\usepackage{inconsolata}

\usepackage{graphicx}

\usepackage{booktabs}
\usepackage{enumitem}
\usepackage{multirow}
\usepackage{amsmath}
\usepackage{amssymb}
\usepackage{algorithm}
\usepackage{algpseudocode}

\title{Reliable Knowledge Acquisition for Continual Learning Agents}

\author{Yixu Huang \\ University of California, Davis \And Jiaqi W. Ma \\ University of Illinois Urbana-Champaign}

\begin{document}
\maketitle
\begin{abstract}
As agents powered by foundation models move into open-ended environments, continual learning from experience becomes essential. For agents built on models with fixed parameters, this learning requires reusing prior task experience in context to improve performance and efficiency on future tasks. Existing memory and experience-learning systems typically focus on either retained facts or procedures, and lack a mechanism for establishing the reliability of claims extracted from the ongoing task stream. Indeed, we find that approximately 12--33\% of entries produced by representative systems harm future task performance when reused. To address these limitations, we propose \textbf{RIKA} (Reliable Knowledge Acquisition), which combines hierarchical extraction of factual, conceptual, procedural, and metacognitive knowledge with Evidence-based Knowledge Admission, which uses source evidence and subsequent experience to govern which claims enter the task agent's context, without additional validation tasks. Across six task streams from Continual Learning Bench and four model backbones, RIKA achieves the highest aggregate score on every backbone. 

\end{abstract}

\section{Introduction}
\label{sec:introduction}

Foundation-model agents increasingly operate in open-ended environments, where they encounter ongoing streams of tasks whose demands cannot be fully anticipated during training~\citep{zhang2025agent,liu2024agentbench,shu2026agentcl,zheng2025lifelongagentbench}. Their experience provides a continuing source of learning (e.g., lessons from failures). For agents built on black-box language models with fixed parameters, this learning takes place through context: knowledge acquired from prior trajectories is supplied to guide subsequent decisions~\citep{shinn2023reflexion}. Continual learning therefore requires organizing past experience into reusable knowledge that can inform future tasks.

A central challenge in reusing experience is \textbf{whether extracted knowledge can be trusted to guide future decisions}.  
Existing memory systems~\citep{chhikara2025mem0,xu2026mem,liu2026simplemem} often lack an explicit mechanism accounting for the reliability of extracted knowledge.
Some experience-learning systems validate extracted knowledge by replaying tasks~\citep{li2026online} or evaluating on held-out tasks~\citep{yang2026skillopt}, both of which require additional task execution that may be unavailable in continual settings, where environment states cannot be revisited or reset. 
Our quantitative evaluation of artifacts produced by representative systems shows that approximately 12--33\% of evaluated atomic claims reduce subsequent task performance when reused. 
A complementary qualitative case study further reveals how such failures arise. Extracted knowledge may rely on information unavailable at reuse time, mismatch the requirements of subsequent tasks, or become outdated as the environment changes (Section~\ref{sec:reliability-diagnosis}).

Beyond reliability, continual learning also requires deciding \textbf{what knowledge to acquire}. Memory systems mostly retain facts and events, while experience-learning systems distill skills, workflows, reflections, and strategic guidance~\citep{wang2024agent,zhang2026agentic,ni2026trace2skill,yang2026skillopt,shinn2023reflexion}. Beyond observations and procedures, agents also need to learn underlying principles and when to apply or revise a method. This motivates a structured approach to learning complementary forms of knowledge and organizing the relationships among them.

To address these two challenges, we propose \textbf{RIKA} (\underline{\textbf{R}}el\underline{\textbf{i}}able \underline{\textbf{K}}nowledge \underline{\textbf{A}}cquisition), a continual learning framework that transforms agents' task experience into structured knowledge for future decisions. To ensure reliable reuse, RIKA introduces \textbf{Evidence-based Knowledge Admission (EKA)}, which draws on sequential hypothesis testing and truth maintenance~\citep{wald1992sequential,doyle1979truth} to admit and revise atomic claims as evidence accumulates from the ongoing task stream. Claims directly established by their source trajectories can be admitted immediately, while those requiring broader support remain pending until sufficient evidence accumulates; subsequent experience can also challenge earlier admissions. This process requires no additional validation tasks. For knowledge extraction, RIKA draws on a knowledge taxonomy from educational psychology~\citep{anderson2001taxonomy} to acquire four complementary types of knowledge: factual, conceptual, procedural, and metacognitive. Its hierarchical extractor grounds conceptual abstractions in factual knowledge and metacognitive abstractions in procedural knowledge, organizing both concrete experience and higher-level understanding for reuse.

Our contributions are threefold:
\begin{itemize}[leftmargin=*, itemsep=2pt, topsep=4pt]
    \item We identify and quantify a reliability problem in experience reuse, finding that approximately 12--33\% of evaluated entries from representative systems harm subsequent task performance.
    \item We propose RIKA, combining evidence-based knowledge admission with hierarchical extraction of factual, conceptual, procedural, and metacognitive knowledge.
    \item We demonstrate RIKA's effectiveness across different task streams and model backbones, achieving the highest aggregate score on all backbones.
\end{itemize}

\begin{figure*}[t]
\centering
\includegraphics[width=\textwidth]{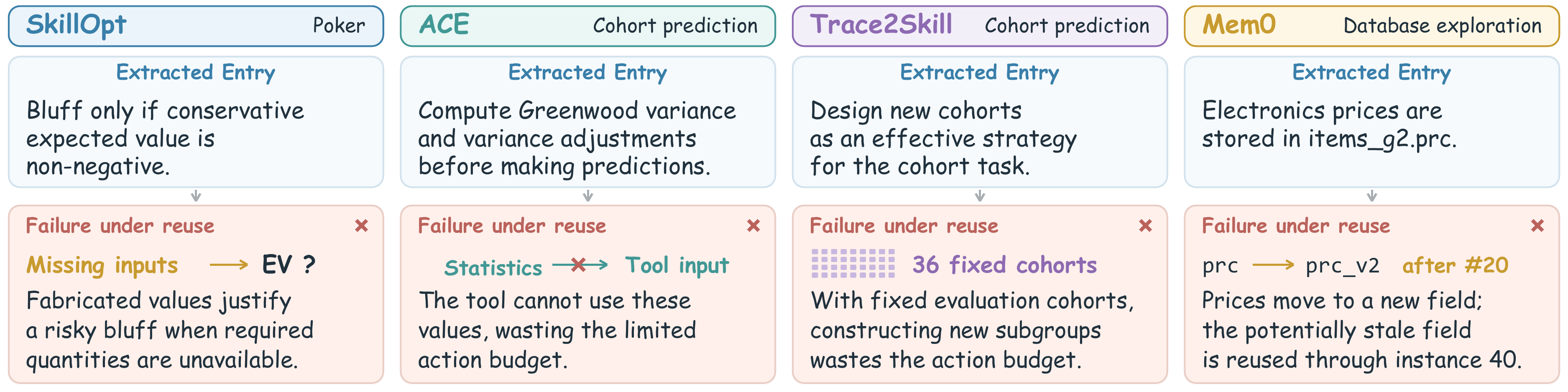}
\caption{\textbf{Extracted knowledge that fails under reuse.}}
\label{fig:unreliable-cases}
\vspace{-1em}
\end{figure*}

\section{Related Work}
\label{sec:related-work}

\subsection{Memory for Agents}

Agent memory systems maintain information from prior interactions for later use. Much of this work is rooted in conversational personal assistants, where memory centers on past events, personal information, and user preferences~\citep{maharana2024evaluating,wu2024longmemeval,hu2026evaluating}. These systems develop memory storage, extraction, and organization mechanisms to retain and retrieve information across interactions~\citep{sumers2023cognitive,zhong2024memorybank,packer2023memgpt,chhikara2025mem0,xu2026mem,liu2026simplemem}. Together, these efforts help agents preserve relevant context across interactions.

\subsection{Learning from Task Experience}

Another line of work extracts reusable knowledge from task experience. Some approaches distill experience into executable skills~\citep{wang2023voyager, ni2026trace2skill}, action workflows~\citep{wang2024agent}, reflections~\citep{shinn2023reflexion}, reasoning insights~\citep{ouyang2026reasoningbank}, and playbooks~\citep{zhang2026agentic}. While several methods validate acquired knowledge before reuse, they rely on replaying tasks or evaluating on held-out tasks~\citep{alzubi2026evoskill,li2026online,yang2026skillopt}, which requires additional task execution that is costly and often infeasible once the environment changes. In contrast, our method assesses knowledge using evidence from the ongoing task stream.

\subsection{Reliability of Agentic Memory}

The reliability of stored knowledge can deteriorate as agents accumulate experience. Recent studies identify errors introduced by repeated consolidation~\citep{zhang2026useful}, stale knowledge under environmental change~\citep{chao2026stale}, and failures when agents consume conflicting memories~\citep{chen2026compliance}. Complementary empirical work examines how memory management affects agents' experience-following behavior~\citep{xiong2026memory}. Beyond characterizing these risks, TrustMem evaluates memory updates for coverage, preservation, and faithfulness, and uses preference-guided reinforcement learning to improve consolidation~\citep{yang2026trustmem}. Our work focuses on assessing the reliability of extracted knowledge using evidence from the ongoing task stream.

\section{Reliability of Knowledge Extraction in Existing Systems}
\label{sec:trust-analysis}

We first diagnose reliability failures in artifacts produced by four representative systems: Mem0~\citep{chhikara2025mem0}, ACE~\citep{zhang2026agentic}, Trace2Skill~\citep{ni2026trace2skill}, and SkillOpt~\citep{yang2026skillopt}. We then introduce EKA to govern knowledge reuse and evaluate it with the SkillOpt and Trace2Skill extractors. All analyses in this section useDeepSeek-V4-Flash~\citep{deepseekai2026deepseekv4} on three task streams (\textsc{DB}, \textsc{Cohort}, and \textsc{Codebase}) from Continual Learning Bench~\citep{asawa2026continual} (Section~\ref{sec:experimental-setup}).

\subsection{How Reliable Is Extracted Knowledge?}
\label{sec:reliability-diagnosis}

\paragraph{Quantitative analysis.}
\label{sec:trust-quantitative}

We examine the reliability of extracted knowledge through a leave-one-entry-out evaluation. Specifically, we remove one knowledge entry at a time and compare downstream task performance with and without that entry. This allows us to identify harmful entries, whose inclusion significantly degrades performance on subsequent tasks.

Since systems package knowledge at different granularities, we decompose every artifact into atomic claims and treat each claim as one entry. Following influence analyses of contextual elements~\citep{nguyen2023incontext,xie2025demoshapley,hu2026skillbrew}, we define the contribution of an entry \(a\) in a frozen artifact \(\mathcal{A}\) as
\[
\Delta(a)
=
\mathbb{E}_{t \in \mathcal{I}(a)}
\left[
\mathrm{score}(t \mid \mathcal{A})
-
\mathrm{score}(t \mid \mathcal{A} \setminus \{a\})
\right],
\]
where \(\mathcal{I}(a)\) contains the future tasks on which the original system would expose \(a\). We classify \(a\) as \emph{harmful} if the 95\% confidence interval of \(\Delta(a)\) lies entirely below zero (Appendix~\ref{sec:appendix-harmful-evaluation}).

\begin{table}[t]
\centering
\small
\setlength{\tabcolsep}{5pt}
\begin{tabular}{lccc}
\toprule
\textbf{Method} &
\textbf{\textsc{DB}} &
\textbf{\textsc{Cohort}} &
\textbf{\textsc{Codebase}} \\
\midrule
Mem0        & 11.7\% & 13.4\% & 12.5\% \\
ACE         & 20.2\% & 18.9\% & 23.1\% \\
Trace2Skill & 32.7\% & 28.8\% & 26.0\% \\
SkillOpt    & 25.9\% & 26.1\% & 28.0\% \\
\bottomrule
\end{tabular}
\caption{
\textbf{Harmful entry rates under leave-one-entry-out evaluation.} Rates are computed over entries within each artifact and averaged equally across artifacts.
}
\vspace{-1.5em}
\label{tab:harmful-entry-rate}
\end{table}

Table~\ref{tab:harmful-entry-rate} shows that approximately 12--33\% of evaluated entries reduce subsequent task performance across the four systems and three task streams. A substantial fraction of retained knowledge therefore hurts future decisions.

\paragraph{Case study.}
\label{sec:trust-case-study}

To understand how such failures arise, we compare claims extracted by these systems with their source trajectories and the requirements of subsequent tasks. Figure~\ref{fig:unreliable-cases} presents four representative cases. Extracted knowledge may rely on information unavailable at reuse time, mismatch the requirements of subsequent tasks, or become outdated as the environment changes.

Together, these findings show that extraction and admission should be treated separately: a plausible claim is not necessarily reliable enough to guide future decisions. This motivates mechanisms for deciding which knowledge to admit and revising those decisions as new evidence accumulates.

\subsection{Evidence-Based Knowledge Admission}
\label{sec:stream-admission}

We introduce \textbf{Evidence-based Knowledge Admission (EKA)} to govern reuse through evidence. EKA distinguishes claims by whether environment feedback in their source trajectories already establishes them. A claim directly verified by such feedback can be admitted immediately. For example, after an agent inspects a database schema, the claim ``in schema version~1, order amounts are stored in \texttt{orders.amount}'' is confirmed by the inspection output. In contrast, a claim that generalizes beyond what was observed remains pending until corroborated by subsequent experience. For example, the rule ``for realized-revenue queries, pending orders do not contribute to revenue'' is inferred from the records seen in a single task, so it is admitted or rejected only after further evidence emerges. The ongoing task stream thus serves as a source of both knowledge and evidence for its reuse. EKA encompasses atomization, claim routing, relation judgment, evidence accumulation, admission and forgetting, and rendering for reuse (Section~\ref{sec:when-to-trust}).

\subsection{Applying EKA to Other Extractors}
\label{sec:admission-transfer}

We evaluate EKA using knowledge extracted by SkillOpt and Trace2Skill, coupling each extractor with EKA and comparing them with their original counterparts. See details in Appendix~\ref{sec:appendix-transfer}.

\begin{table}[t]
\centering
\small
\setlength{\tabcolsep}{5pt}
\begin{tabular}{lccc}
\toprule
\textbf{Method} & \textbf{\textsc{DB}} & \textbf{\textsc{Cohort}} & \textbf{\textsc{Codebase}} \\
\midrule
SkillOpt & 0.147 & -0.062 & 0.567 \\
SkillOpt + EKA & 0.192 & -0.045 & 0.573 \\
Trace2Skill & 0.158 & -0.053 & 0.538 \\
Trace2Skill + EKA & 0.224 & -0.043 & 0.570 \\
\bottomrule
\end{tabular}
\caption{\textbf{Task performance of SkillOpt and Trace2Skill with and without EKA.} Scores are averaged over five runs. Each + EKA variant retains the system's extractor and applies EKA thereafter.}
\label{tab:admission-transfer}
\end{table}

Table~\ref{tab:admission-transfer} demonstrates that EKA improves both systems on \textsc{DB} and \textsc{Cohort}, supporting the effectiveness of the mechanism with different extractors.

\section{Method}
\label{sec:method}

\begin{figure*}[t]
    \centering
    \includegraphics[width=\textwidth]{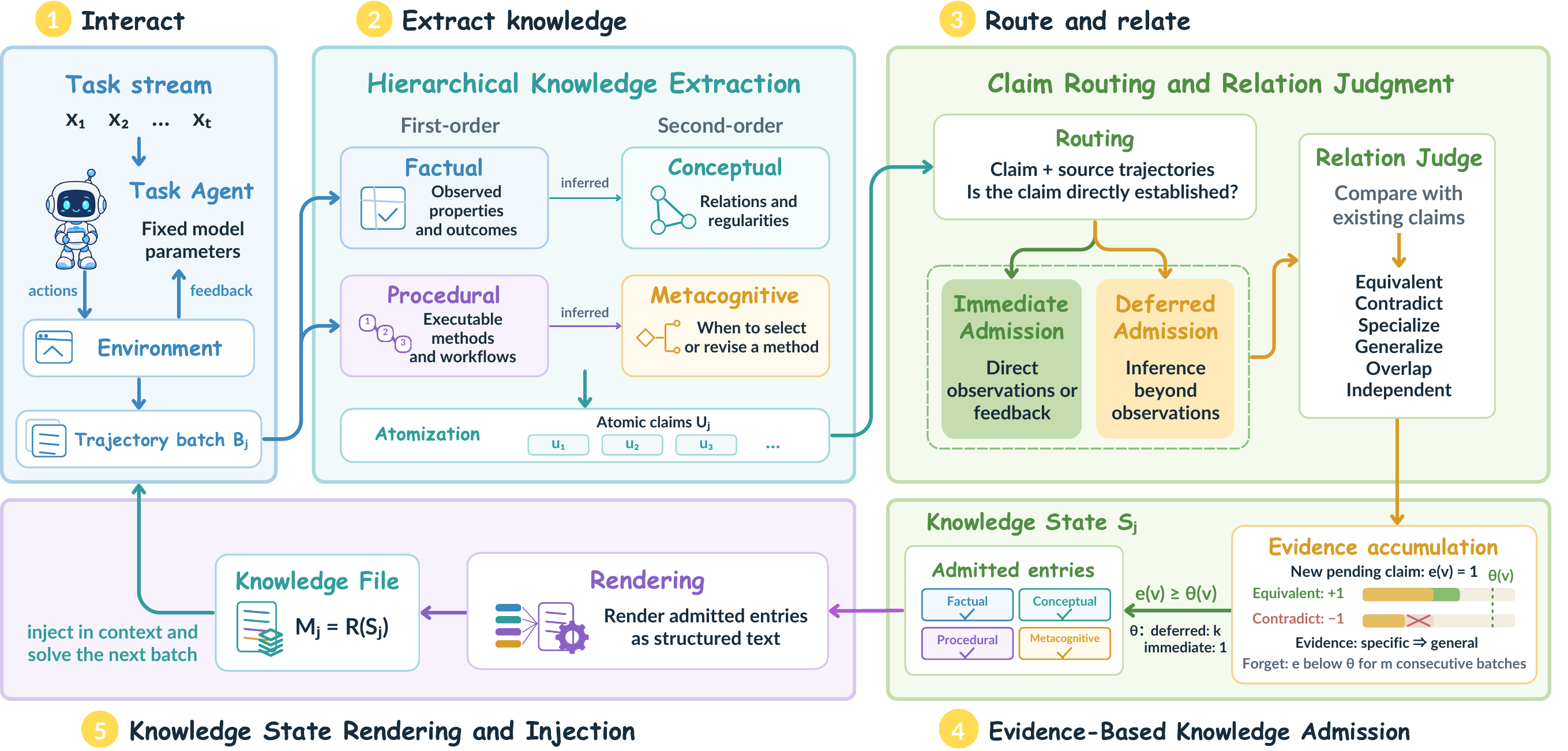}
    \caption{\textbf{RIKA: hierarchical extraction with evidence-based knowledge admission.}
    }
    \vspace{-1em}
    \label{fig:our-method}
\end{figure*}

RIKA combines hierarchical knowledge extraction with EKA (Figure~\ref{fig:our-method}). The extractor acquires four complementary types of knowledge; EKA processes its outputs through atomization, routing, relation judgment, evidence accumulation, admission and forgetting, and rendering for reuse.

\subsection{Setup: Continual Learning from Task Streams}

We consider a task agent that interacts with an environment over a stream of tasks
\[
\mathcal{T}=\{x_1,\ldots,x_T\}.
\]

Solving task $x_t$ produces a trajectory $\tau_t$ consisting of the agent's actions, environment observations, and feedback. We focus on agents built on black-box models whose parameters remain fixed; continual learning therefore occurs through reusable knowledge acquired from past trajectories and supplied in context for future tasks. The goal is to improve task performance and execution efficiency as experience accumulates.

We define the \(j\)-th batch as
\[
B_j=\{\tau_{(j-1)n+1},\ldots,\tau_{jn}\}.
\]
After completing \(B_j\), the system updates its persistent knowledge state from \(S_{j-1}\) to \(S_j\) and produces a textual knowledge file \(M_j\) for solving the subsequent batch. The remaining question is how this knowledge state should be constructed and updated from task experience.

Throughout, a \emph{claim} denotes a self-contained unit of reusable knowledge, whether newly extracted or stored in the persistent state. Equivalent extracted claims are consolidated into a canonical claim, stored with its metadata, provenance, evidence, relations, and admission route. We use \(u\) for a newly extracted claim and \(v\) for a stored claim. Admission is reassessed after each batch.

\subsection{Knowledge Representation}
\label{sec:what-to-learn}

Inspired by a knowledge taxonomy from educational psychology~\citep{anderson2001taxonomy}, we represent task experience using four complementary knowledge types, illustrated below with examples from the \textsc{DB} task. \textbf{Factual knowledge} captures task-relevant properties of entities, states, and outcomes (e.g., ``\texttt{fdbk\_g1.ts} stores Unix timestamps in milliseconds''). \textbf{Conceptual knowledge} captures relations or regularities that organize multiple facts (e.g., ``timestamp encodings differ across feedback tables: milliseconds, seconds, or ISO date strings''). \textbf{Procedural knowledge} specifies executable methods or workflows (e.g., ``to filter by a top-level category, use \texttt{cat\_lvl = 1 AND cat\_nm = '<Category>'}''). \textbf{Metacognitive knowledge} concerns when to apply, select, or revise a method (e.g., ``when a category filter returns unexpected results, verify the hierarchy level before revising the query'').

\subsection{Hierarchical Knowledge Extraction}

The four knowledge types differ in their level of abstraction~\citep{anderson2001taxonomy}. Factual and procedural knowledge describe concrete observations and task-solving operations~\citep{anderson1982acquisition}, which can be directly grounded in environment feedback and agent trajectories. In contrast, conceptual and metacognitive knowledge capture higher-level regularities and strategic judgment~\citep{nelson1990metamemory}, which typically require further abstraction over factual and procedural knowledge, respectively. We therefore organize knowledge extraction into two stages: first extracting factual and procedural knowledge, and then deriving abstract conceptual and metacognitive knowledge from them, respectively.

For each batch \(B_j\), since each type requires a distinct extraction objective, we instantiate four type-specific extraction subagents that operate without access to stored claims. We first extract first-order factual and procedural knowledge grounded in the environment feedback in the trajectories:
\[
F_j=\mathcal{E}_{F}(B_j),
\qquad
P_j=\mathcal{E}_{P}(B_j).
\]
Second-order knowledge is then extracted conditioned on both the trajectories and the corresponding first-order knowledge:
\[
C_j=\mathcal{E}_{C}(B_j,F_j),
\qquad
G_j=\mathcal{E}_{M}(B_j,P_j),
\]
where \(C_j\) and \(G_j\) denote conceptual and metacognitive knowledge, respectively. Each extractor processes the trajectories in the batch jointly.

\subsection{Evidence-Based Knowledge Admission (EKA)}
\label{sec:when-to-trust}

EKA evaluates claims from the ongoing task stream. Inspired by sequential hypothesis testing~\citep{wald1992sequential} and truth maintenance~\citep{doyle1979truth}, we keep eligible but insufficiently corroborated claims pending and update their evidence as subsequent trajectories provide support or opposition. EKA proceeds through stages below.

\subsubsection{Atomization}
An extractor output may bundle multiple claims within a single item. Admitting or rejecting such an item as a whole using EKA might let the unsupported part into context or discard the supported part. 
Inspired by atomic decomposition~\citep{min2023factscore}, we use an atomizer across all four knowledge types, splitting claims only when their components are self-contained and independent. Each resulting atom must be grounded in the environment feedback relevant to its own content, either directly or, for second-order knowledge, through its first-order claims. The atomic claims serve as the basic units for relation judgment and evidence accumulation. We denote those extracted from batch \(B_j\) by \(U_j=\{u_1,\ldots,u_{|U_j|}\}\). Appendix~\ref{sec:appendix-prompts} reports technical details including prompts and output schemas.

\subsubsection{Claim Routing and Relation Judgment}
\label{sec:relation} 

\paragraph{Admission routing.}
A routing agent examines each atomic claim \(u\), its cited environment-feedback spans, and the explanation linking them. A claim is \emph{directly} grounded if the cited feedback itself states the claim under its stated scope, and \emph{indirectly} grounded if it follows from such feedback through inference or abstraction, possibly via its first-order parent claims; claims with neither are excluded. The grounding type determines the admission route. Directly grounded claims follow the \textbf{immediate admission} path, since the feedback already establishes them. Indirectly grounded claims follow the \textbf{deferred admission} path, since their full scope still requires corroboration, for example because of causal attribution or conditions beyond those observed. We denote the route by
\[
z(u)\in\{\textsc{Immediate},\textsc{Deferred}\}.
\]
The route depends on whether the evidence covers a claim's scope, not on its knowledge type. A factual claim such as ``schema versions use \texttt{completed} as the finalized label'' is deferred, since an observation from version~1 does not cover its cross-version scope. Conversely, a conceptual claim can be admitted immediately when it only relates directly observed facts within their observed scope, such as ``timestamp encodings differ across \texttt{fdbk\_g1}, \texttt{fdbk\_g2}, and \texttt{fdbk\_g3}''. Routing decisions are recorded when extracted claims are integrated into the persistent state.

\paragraph{Relation judgment.} Inspired by natural-logic approaches to textual inference~\citep{maccartney2007natural}, we compare each new extracted claim with existing claims. For a new extracted claim \(u\) and an existing claim \(v\), the relation judge predicts one or more material logical relations \(r(u,v) \subseteq \mathcal{R}\) between \(u\) and \(v\).
We define \(\mathcal{R}\) to contain \textsc{Equivalent}, \textsc{Specialize}, \textsc{Generalize}, \textsc{Contradict}, \textsc{Overlap}, and \textsc{Independent}. The relations are defined as follows:
\begin{itemize}[leftmargin=*, itemsep=2pt, topsep=4pt]
    \item \textsc{Equivalent}: \(u\) and \(v\) entail each other under the same scope.
    \item \textsc{Specialize}: The new claim \(u\) entails the broader stored claim \(v\).
    \item \textsc{Generalize}: The stored claim \(v\) entails the broader new claim \(u\).
    \item \textsc{Contradict}: The claims cover overlapping scope but cannot both hold.
    \item \textsc{Overlap}: Shared content without entailment or contradiction.
    \item \textsc{Independent}: No logical relationship.
\end{itemize}
These judgments determine how subsequent experience supports or challenges stored claims.

\begin{table*}[t]
\centering
\tiny
\setlength{\tabcolsep}{2.2pt}
\resizebox{\textwidth}{!}{%
\begin{tabular}{ll*{14}{c}}
\toprule
\multirow{2}{*}{\textbf{Model}} & \multirow{2}{*}{\textbf{Method}}
& \multicolumn{2}{c}{\textbf{\textsc{DB}}}
& \multicolumn{2}{c}{\textbf{\textsc{Poker}}}
& \multicolumn{2}{c}{\textbf{\textsc{Cohort}}}
& \multicolumn{2}{c}{\textbf{\textsc{BSM}}}
& \multicolumn{2}{c}{\textbf{\textsc{Sales}}}
& \multicolumn{2}{c}{\textbf{\textsc{Codebase}}}
& \multicolumn{2}{c}{\textbf{Agg.}} \\
\cmidrule(lr){3-4}\cmidrule(lr){5-6}\cmidrule(lr){7-8}\cmidrule(lr){9-10}\cmidrule(lr){11-12}\cmidrule(lr){13-14}\cmidrule(lr){15-16}
& & Score & Gain & Score & Gain & Score & Gain & Score & Gain & Score & Gain & Score & Gain & Score & Gain \\
\midrule

\multirow{6}{*}{GPT-5.4}
& \textbf{RIKA}
& 0.351 & +0.213 & \textbf{1.247} & \textbf{+0.134} & -0.020 & -0.013 & \textbf{0.579} & \textbf{+0.360} & \textbf{0.808} & \textbf{+0.342} & 0.551 & +0.053 & \textbf{0.232} & \textbf{+0.232} \\
& ICL
& 0.347 & +0.209 & 0.798 & -0.315 & -0.028 & -0.021 & 0.513 & +0.294 & 0.769 & +0.303 & 0.534 & +0.036 & 0.183 & +0.183 \\
& Mem0
& \textbf{0.431} & \textbf{+0.293} & 0.612 & -0.501 & -0.034 & -0.027 & 0.375 & +0.156 & 0.651 & +0.185 & 0.584 & +0.086 & 0.140 & +0.140 \\
& ACE
& 0.196 & +0.058 & 1.196 & +0.083 & \textbf{-0.004} & \textbf{+0.003} & 0.220 & +0.001 & 0.510 & +0.044 & \textbf{0.609} & \textbf{+0.111} & 0.067 & +0.067 \\
& Trace2Skill
& 0.138 & +0.000 & 1.115 & +0.002 & -0.032 & -0.025 & 0.318 & +0.099 & 0.689 & +0.223 & 0.513 & +0.015 & 0.071 & +0.071 \\
& SkillOpt
& 0.189 & +0.051 & 0.971 & -0.142 & -0.024 & -0.017 & 0.443 & +0.224 & 0.566 & +0.100 & 0.607 & +0.109 & 0.105 & +0.105 \\

\midrule
\multirow{6}{*}{DeepSeek-V4-Flash}
& \textbf{RIKA}
& \textbf{0.281} & \textbf{+0.221} & \textbf{1.141} & \textbf{+0.489} & -0.042 & +0.076 & 0.234 & +0.015 & \textbf{0.847} & \textbf{+0.333} & \textbf{0.575} & \textbf{+0.063} & \textbf{0.141} & \textbf{+0.233} \\
& ICL
& 0.256 & +0.196 & 0.755 & +0.103 & -0.084 & +0.034 & \textbf{0.518} & \textbf{+0.299} & 0.639 & +0.125 & 0.548 & +0.036 & 0.074 & +0.176 \\
& Mem0
& 0.274 & +0.214 & 0.732 & +0.080 & \textbf{-0.038} & \textbf{+0.080} & 0.348 & +0.129 & 0.522 & +0.008 & 0.547 & +0.035 & 0.049 & +0.129 \\
& ACE
& 0.135 & +0.075 & 0.637 & -0.015 & -0.059 & +0.059 & 0.222 & +0.003 & 0.563 & +0.049 & 0.529 & +0.017 & -0.020 & +0.071 \\
& Trace2Skill
& 0.158 & +0.098 & 0.858 & +0.206 & -0.053 & +0.065 & 0.293 & +0.074 & 0.602 & +0.088 & 0.538 & +0.026 & 0.025 & +0.115 \\
& SkillOpt
& 0.147 & +0.087 & 0.435 & -0.217 & -0.062 & +0.056 & 0.218 & -0.001 & 0.658 & +0.144 & 0.567 & +0.055 & 0.017 & +0.113 \\

\midrule
\multirow{6}{*}{MiniMax-M3}
& \textbf{RIKA}
& \textbf{0.327} & \textbf{+0.191} & \textbf{0.623} & \textbf{+0.758} & \textbf{0.012} & \textbf{+0.017} & 0.257 & +0.037 & 0.758 & +0.561 & \textbf{0.554} & \textbf{+0.181} & \textbf{0.163} & \textbf{+0.239} \\
& ICL
& 0.284 & +0.148 & 0.407 & +0.542 & -0.020 & -0.015 & \textbf{0.337} & \textbf{+0.117} & 0.535 & +0.338 & 0.483 & +0.110 & 0.043 & +0.147 \\
& Mem0
& 0.303 & +0.167 & 0.235 & +0.370 & 0.003 & +0.008 & 0.289 & +0.069 & 0.501 & +0.304 & 0.402 & +0.029 & 0.018 & +0.132 \\
& ACE
& 0.189 & +0.053 & 0.509 & +0.644 & -0.026 & -0.021 & 0.230 & +0.010 & \textbf{0.772} & \textbf{+0.575} & 0.513 & +0.140 & 0.082 & +0.159 \\
& Trace2Skill
& 0.175 & +0.039 & 0.612 & +0.747 & -0.031 & -0.026 & 0.219 & -0.001 & 0.638 & +0.441 & 0.533 & +0.160 & 0.039 & +0.128 \\
& SkillOpt
& 0.180 & +0.044 & 0.559 & +0.694 & -0.028 & -0.023 & 0.234 & +0.014 & 0.672 & +0.475 & 0.517 & +0.144 & 0.050 & +0.137 \\

\midrule
\multirow{6}{*}{Gemini-3-Flash}
& \textbf{RIKA}
& \textbf{0.387} & \textbf{+0.298} & \textbf{0.952} & \textbf{-0.688} & -0.058 & +0.001 & \textbf{0.434} & \textbf{+0.215} & 0.628 & +0.189 & 0.614 & +0.207 & \textbf{0.130} & \textbf{+0.201} \\
& ICL
& 0.376 & +0.287 & 0.790 & -0.850 & \textbf{-0.034} & \textbf{+0.025} & 0.367 & +0.148 & \textbf{0.707} & \textbf{+0.268} & 0.390 & -0.017 & 0.084 & +0.160 \\
& Mem0
& 0.355 & +0.266 & 0.863 & -0.777 & -0.038 & +0.021 & 0.334 & +0.115 & 0.544 & +0.105 & 0.583 & +0.176 & 0.084 & +0.153 \\
& ACE
& 0.239 & +0.150 & 0.452 & -1.188 & -0.074 & -0.015 & 0.378 & +0.159 & 0.591 & +0.152 & 0.517 & +0.110 & 0.020 & +0.101 \\
& Trace2Skill
& 0.098 & +0.009 & 0.539 & -1.101 & -0.095 & -0.036 & 0.395 & +0.176 & 0.538 & +0.099 & \textbf{0.622} & \textbf{+0.215} & -0.005 & +0.079 \\
& SkillOpt
& 0.103 & +0.014 & 0.588 & -1.052 & -0.087 & -0.028 & 0.385 & +0.166 & 0.559 & +0.120 & 0.605 & +0.198 & 0.004 & +0.086 \\

\bottomrule
\end{tabular}%
}
\caption{
\textbf{Continual-learning performance across streams and backbones.}
\textsc{Score}: stream-level score; \textsc{Gain}: improvement over the shared stateless control for the same backbone and task; \textsc{Agg.}: averages of normalized scores and gains across six streams. See Appendices~\ref{sec:appendix-setup} and~\ref{sec:appendix-main-ci} for setup details and 95\% confidence intervals.}
\label{tab:main-results}
\end{table*}

\subsubsection{Evidence Accumulation, Admission, and Forgetting}
\label{sec:admission}

After each batch, we update the evidence for all active claims and reassess their admission, including claims admitted in earlier batches. Full update rules are provided in Appendix~\ref{sec:appendix-algorithm}.

\paragraph{Relation-guided evidence accumulation.}
Each stored claim \(v\) maintains a cumulative net evidence count \(e_j(v)\). Let \(c_j^{+}(v)\) and \(c_j^{-}(v)\) denote the supporting and opposing evidence counts from batch \(B_j\). We update
\[
e_j(v)=e_{j-1}(v)+c_j^{+}(v)-c_j^{-}(v),
\]
with the count initialized to zero for a new claim.

Fresh trajectories support a claim when they ground an equivalent or more specific claim, and oppose it when they ground a contradictory claim. Equivalent claims are merged into a entry. Entailment edges form a directed acyclic graph from specific claims to general claims: support propagates to reachable generalizations, but not in the reverse direction. \textsc{Overlap} and \textsc{Independent} relations transfer no evidence.

\paragraph{Admission.}
We use the admission threshold \(\theta(v)\), set to \(1\) for immediate claims and \(k\) for deferred claims. Claims directly established by their source trajectories thus require no additional corroboration, whereas deferred claims require sufficient accumulated net evidence. After each batch, a claim is admitted if its updated evidence count satisfies \(e(v)\geq\theta(v)\) and remains pending otherwise, regardless of its previous admission status. A change in admission does not reset its evidence history.

\paragraph{Forgetting.}
To keep the state bounded, a claim is forgotten if its evidence stays below its admission threshold for \(m\) consecutive batches; reaching the threshold resets this window. The window tolerates temporary lack of support, during which the claim remains pending in the state.

\subsection{Knowledge State Rendering and Injection}
\label{sec:render}

After all updates for batch \(B_j\) are complete, we render the admitted claims in the resulting state \(S_j\), i.e., those satisfying \(e(v)\geq\theta(v)\), into a structured textual knowledge file
\[
M_j = \mathcal{R}(S_j).
\]
Motivated by findings that repeated memory updates can degrade stored knowledge~\citep{zhang2026useful,yang2026trustmem}, we reconstruct the file from the current state rather than incrementally rewriting \(M_{j-1}\).

Finally, the complete rendered file \(M_j\) is supplied in context to the task agent while it solves the subsequent batch (Appendix~\ref{sec:appendix-rendering}). The resulting trajectories provide the next source of experience, closing the continual learning loop.

\section{Experiments}
\label{sec:experiments}

We evaluate whether RIKA enables a task agent to continually improve from its task-solving experience across task streams and model backbones.

\subsection{Experimental Setup}
\label{sec:experimental-setup}

\paragraph{Benchmark.}
We utilize Continual Learning Bench, evaluating on six heterogeneous  task streams: \textsc{db}, \textsc{poker}, \textsc{cohort}, \textsc{bsm}, \textsc{sales}, and \textsc{codebase}~\citep{asawa2026continual}. See task details and scoring in Appendix~\ref{sec:appendix-streams} and Appendix~\ref{sec:appendix-scoring}.

\paragraph{Baselines.}
We choose five representative baselines: ICL~\citep{asawa2026continual}, which directly reuses past trajectories in context; Mem0~\citep{chhikara2025mem0}, a general-purpose memory system; ACE~\citep{zhang2026agentic}, which maintains an evolving playbook; and Trace2Skill~\citep{ni2026trace2skill} and SkillOpt~\citep{yang2026skillopt}, which distill trajectories into reusable skills.

\paragraph{Model backbones.}
We evaluate all methods on four backbones: GPT-5.4~\citep{openai2026gpt54}, DeepSeek-V4-Flash~\citep{deepseekai2026deepseekv4}, MiniMax-M3~\citep{lai2026minimax}, and Gemini-3-Flash~\citep{googledeepmind2025gemini3flash}.
Appendix~\ref{sec:appendix-model-config} reports model versions and the shared configuration, context management, and adaptation settings.

\paragraph{Evaluation.}
By default, RIKA uses batch size \(n=5\) (\(n=4\) for \textsc{Sales}, which has only 12 instances), deferred admission threshold \(k=3\), and forgetting window \(m=3\). All results are averaged over five runs (Appendix~\ref{sec:appendix-runs}).

For each stream in Table~\ref{tab:main-results}, we report the mean task reward (\textsc{Score}) and its improvement over a no-learning reference (\textsc{Gain}). The reference is the mean of five stateless runs for the same backbone and task, shared across all methods. Since reward scales differ across tasks, we also report normalized aggregates (\textsc{Agg.}). For learning dynamics, we plot cumulative gain over the stateless control (defined in Appendix~\ref{sec:appendix-stateless}). Curves show the mean over five runs with 95\% confidence intervals.

\begin{figure*}[!t]
    \centering
    \begin{minipage}[b]{0.36\textwidth}
        \centering
        \includegraphics[width=\linewidth]{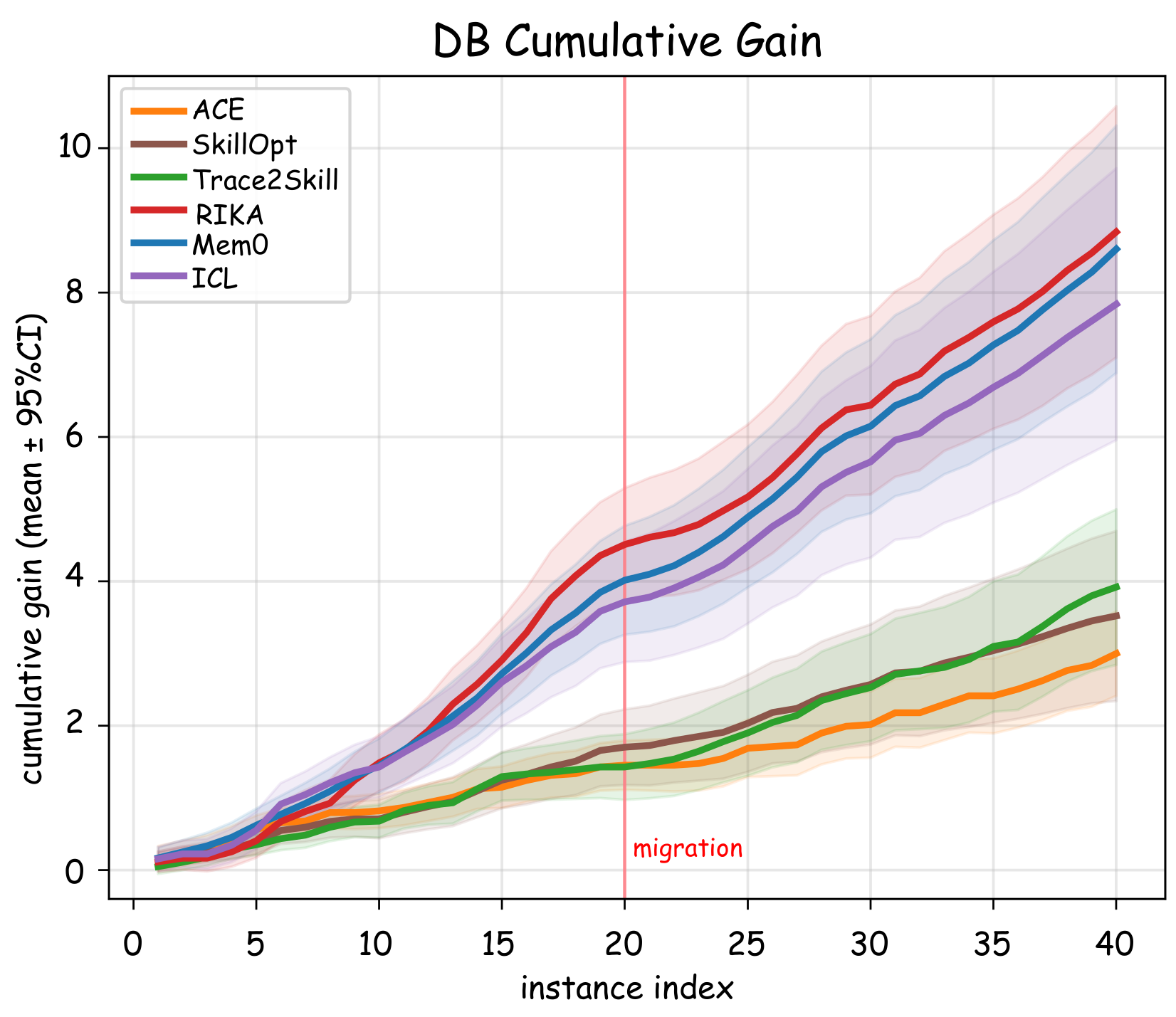}
        \\[2pt]
        \vspace{-0.5em}
        {\small (a) \textsc{DB}}
    \end{minipage}\hfill
    \begin{minipage}[b]{0.57\textwidth}
        \centering
        \includegraphics[width=\linewidth]{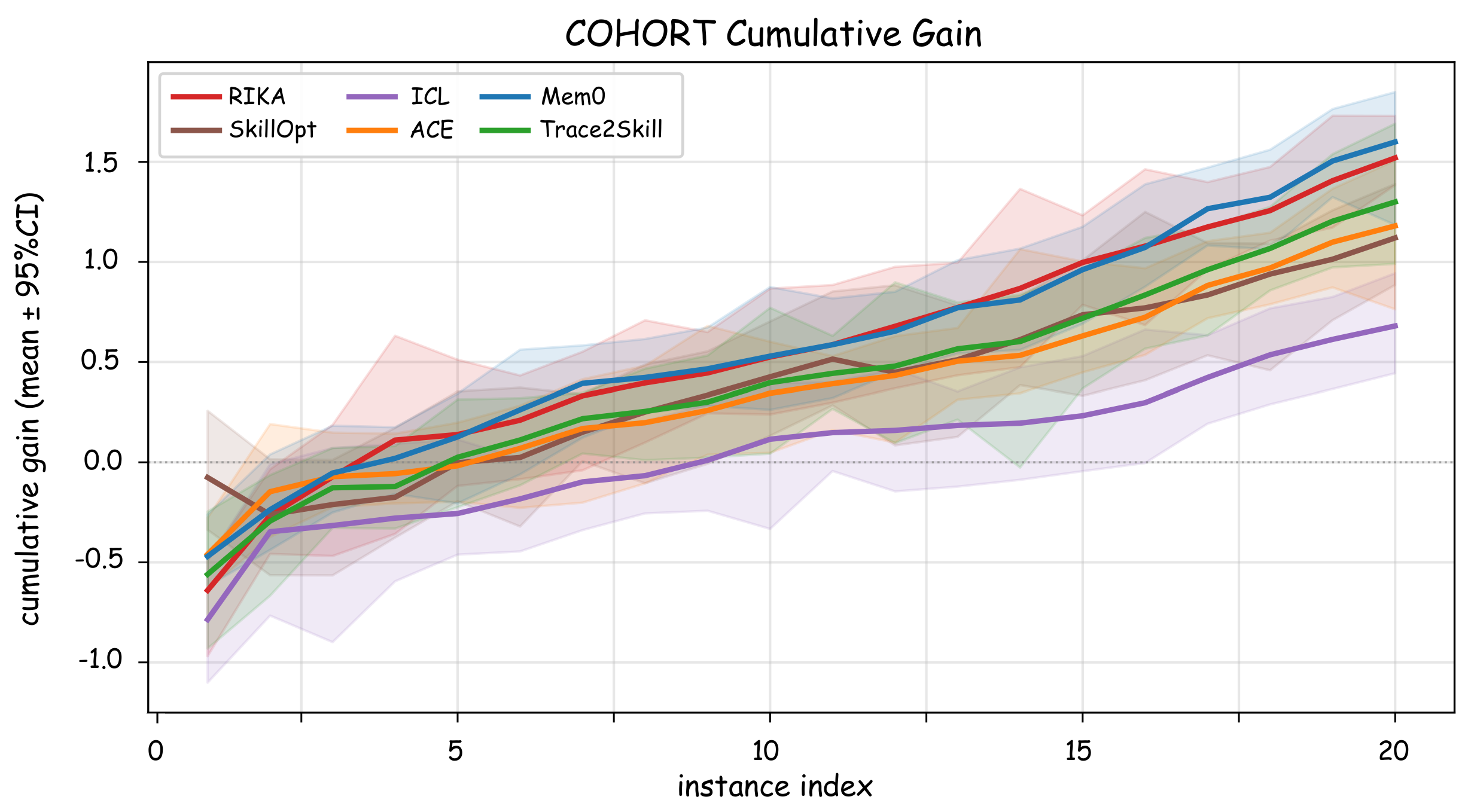}
        \\[2pt]
        \vspace{-0.5em}
        {\small (b) \textsc{Cohort}}
    \end{minipage}
    \caption{\textbf{Cumulative gain on \textsc{DB} and \textsc{Cohort} with DeepSeek-V4-Flash.} Curves show mean gain relative to the stateless control shared across methods within each task; shading indicates 95\% confidence intervals. The vertical line marks the \textsc{DB} migration at instance 20.}
    \label{fig:learning-dynamics}
    \vspace{-1em}
\end{figure*}

\subsection{Main Results}
\label{sec:main-results}

Table~\ref{tab:main-results} presents the results across all six task streams and four model backbones.

\paragraph{Consistent gains across backbones.}
RIKA achieves the highest aggregate score on all backbones, exceeding the strongest baseline on each by $0.046$--$0.081$. For instance, it improves the aggregate score from $0.183$ to $0.232$ on GPT-5.4. Its aggregate gain over the stateless agent is positive on all backbones, ranging from $+0.201$ to $+0.239$.

\paragraph{Broad improvement across task streams.}
RIKA achieves the best result in $14$ of $24$ model--task combinations. The remaining wins are distributed among task-specific baselines, particularly on \textsc{Cohort}, \textsc{BSM}, and \textsc{Codebase}, showing that RIKA's advantage is broad but not uniform.

\paragraph{Structured knowledge outperforms direct experience reuse.}
RIKA outperforms ICL in aggregate on every backbone by $0.046$--$0.120$, suggesting that structured, selectively admitted claims enable more effective reuse than directly exposing the agent to past trajectories.

\section{Analysis}
\label{sec:analysis}

We examine RIKA's learning dynamics, the contributions of its main components, and its sensitivity to batch size and admission threshold.
All ablations use DeepSeek-V4-Flash on three representative task streams: \textsc{DB}, \textsc{Cohort}, and \textsc{Codebase}. Additional ablations are provided in Appendix~\ref{sec:appendix-additional-analysis}.

\subsection{Dynamics}
\label{sec:analysis-dynamics}

Figure~\ref{fig:learning-dynamics} shows cumulative gain relative to the stateless DeepSeek-V4-Flash agent on \textsc{DB} and \textsc{Cohort}. On \textsc{DB}, RIKA establishes an early lead over the baselines and maintains it after the migration event at instance 20, ending at a cumulative gain consistent with the stream-level gain in Table~\ref{tab:main-results}. On \textsc{Cohort}, its cumulative gain moves from negative to positive and finishes above most baselines but below Mem0, consistent with their respective stream-level gains of $0.076$ and $0.080$. These trends illustrate RIKA's gains over time, while showing that its relative advantage varies across task streams.

\subsection{Ablations on Admission Routing}
\label{sec:analysis-routing}

We compare RIKA's admission mechanism with: Imm. Only, which admits every claim immediately; Def. Only, which defers every claim; and Random Router, which assigns claims randomly.

\begin{table}[t]
\centering
\small
\setlength{\tabcolsep}{5pt}
\begin{tabular}{lccc}
\toprule
\textbf{\textsc{Method}} & \textbf{\textsc{DB}} & \textbf{\textsc{Cohort}} & \textbf{\textsc{Codebase}} \\
\midrule
\textbf{RIKA} & \textbf{0.281} & \textbf{-0.042} & \textbf{0.575} \\
Imm. Only & 0.225 & -0.089 & 0.540 \\
Def. Only & 0.265 & -0.070 & 0.557 \\
Random Router & 0.249 & -0.084 & 0.549 \\
\bottomrule
\end{tabular}
\caption{\textbf{Admission-routing ablations.}}
\vspace{-1em}
\label{tab:routing-ablation}
\end{table}

Table~\ref{tab:routing-ablation} reports stream-level scores only; Appendix Table~\ref{tab:routing-ablation-full} provides the complete results. Relative to RIKA, Imm. Only causes the largest score degradation ($0.047$--$0.056$ on DB and COHORT). Def. Only performs better but still trails the full method, suggesting that conservative admission limits premature reuse while delaying directly supported knowledge. Random Router also underperforms on every stream, indicating that the improvements depend on evidence-aware routing.

\subsection{Ablations on Hierarchical Knowledge Extraction}
\label{sec:analysis-extraction}

We compare our hierarchical extractor with Generic 1-type, which uses one undifferentiated extractor; Flat 4-type, which retains all knowledge types but removes first- to second-order conditioning; and variants that remove each type individually.

\begin{table}[t]
\centering
\small
\setlength{\tabcolsep}{5pt}
\begin{tabular}{lccc}
\toprule
\textbf{Method} & \textbf{\textsc{DB}} & \textbf{\textsc{Cohort}} & \textbf{\textsc{Codebase}} \\
\midrule
\textbf{RIKA} & \textbf{0.281} & \textbf{-0.042} & \textbf{0.575} \\
Generic 1-type & 0.213 & -0.087 & 0.541 \\
Flat 4-type & 0.261 & -0.057 & 0.563 \\
$-$Factual & 0.253 & -0.060 & 0.547 \\
$-$Conceptual & 0.265 & -0.065 & 0.553 \\
$-$Procedural & 0.270 & -0.073 & 0.549 \\
$-$Metacognitive & 0.274 & -0.068 & 0.556 \\
\bottomrule
\end{tabular}
\vspace{-0.5em}
\caption{\textbf{Ablations of knowledge types and hierarchical extraction.}}
\label{tab:knowledge-ablation}
\end{table}

Table~\ref{tab:knowledge-ablation} reports stream-level scores only; Appendix Table~\ref{tab:knowledge-ablation-full} provides the complete results. Relative to RIKA, \textsc{Generic 1-type} causes the largest score degradation ($0.045$--$0.068$ on DB and COHORT), while \textsc{Flat 4-type} also consistently underperforms, supporting the importance of hierarchical conditioning. Removing any individual type reduces performance, with the largest drops from removing factual knowledge on \textsc{DB} and procedural knowledge on \textsc{Cohort}.

\section{Conclusion}
\label{sec:conclusion}

We find that approximately 12--33\% of entries retained by existing experience-reuse systems harm subsequent task performance. To address this, we propose RIKA, which combines hierarchical extraction of four knowledge types with EKA, admitting claims as evidence accumulates from the task stream without additional validation tasks. RIKA achieves the highest aggregate score on every backbone across six task streams, and EKA also improves SkillOpt and Trace2Skill.

\section*{Limitations}

Our empirical study focuses on six text-based, tool-mediated task streams and four model backbones. Although the streams span heterogeneous domains and interaction patterns, the results may not directly transfer to more diverse settings, where experience also includes perceptual or physical feedback. In addition, we study continual learning through an external knowledge artifact while keeping the underlying model parameters fixed. How hierarchical extraction and evidence-based admission interact with parameter-updating approaches remains an open direction.

\section*{Ethical Considerations}

Agent trajectories may contain personal information that could persist in the knowledge file, so deployment on real user interactions would require filtering or anonymizing trajectories first; our experiments use benchmarks without real personal data. Finally, our evaluation covers six text-based, tool-mediated task streams and four model backbones, and it does not show that RIKA is safe for high-stakes domains such as medicine, law, or finance, where domain-specific validation and human oversight would be required.




\bibliography{custom}

\appendix

\setcounter{topnumber}{5}
\setcounter{bottomnumber}{3}
\setcounter{totalnumber}{8}
\setcounter{dbltopnumber}{4}
\renewcommand{\topfraction}{0.95}
\renewcommand{\bottomfraction}{0.9}
\renewcommand{\textfraction}{0.05}
\renewcommand{\floatpagefraction}{0.8}
\renewcommand{\dbltopfraction}{0.95}
\renewcommand{\dblfloatpagefraction}{0.8}

\section*{Appendix}

\section{Protocol for Harmful-Entry Evaluation}
\label{sec:appendix-harmful-evaluation}

This section details the artifact-freezing schedule, entry-use conditions, and repeated paired evaluations used in Section~\ref{sec:trust-quantitative}. We use DeepSeek-V4-Flash throughout this diagnostic evaluation. Each comparison pairs the complete artifact with a version that removes only the target entry.

\subsection{Artifact Freezing and Evaluation Windows}
\label{sec:appendix-harmful-windows}

For each method and task stream, we evaluate three frozen artifacts. We freeze each artifact at the prescribed point in its task stream and evaluate its entries on subsequent instances. All instance indices below are one-based and refer to positions in the ordered stream for the corresponding run.
\begin{itemize}[leftmargin=*, itemsep=2pt, topsep=4pt]
    \item \textbf{\textsc{DB}:} we freeze the artifact immediately after instance 30 and evaluate its entries on the final 10 instances (31--40). The schema migration occurs between instances 20 and 21, before the artifact is frozen, so the evaluation window lies entirely within the post-migration stage.
    \item \textbf{\textsc{Cohort}:} we freeze the artifact after instance 10, halfway through the 20-instance stream, and evaluate its entries on instances 11--20.
    \item \textbf{\textsc{Codebase}:} we freeze the artifact after instance 9 (the Tablib stage) and evaluate its entries on instances 10--19 (the Tenacity stage).
\end{itemize}
Each evaluated artifact therefore contains only experience available at its freezing point, and every evaluation instance occurs strictly after that point. The artifact remains fixed throughout its evaluation window and across all five paired evaluation seeds for each entry.

\subsection{Entry-Level Reuse Conditions}
\label{sec:appendix-harmful-reuse}

To ensure consistent entry granularity across systems, we first decompose every frozen artifact from every evaluated system into atomic claims. Each evaluation entry $a$ denotes one atomic claim, and $\mathcal{A}$ denotes the complete collection of claims for that artifact. This normalization gives all systems the same unit of comparison, irrespective of how their original artifacts group knowledge into entries. We evaluate each claim by removing only that claim while retaining all other claims in the artifact. For retrieval-based systems, we include an instance in $\mathcal{I}(a)$ only if the system's retrieval mechanism would select $a$ for that instance from the complete frozen artifact. The original system's exposure pattern determines the eligible task set for each claim. Entries that are never retrieved during the evaluation window, and hence have $\mathcal{I}(a)=\varnothing$, are excluded from both evaluation and the denominator of the harmful-entry rate. For systems that supply the entire artifact as a knowledge file, all entries are exposed at every instance, so all instances in the corresponding evaluation window are eligible.

For each eligible instance, we first obtain the knowledge context from the complete frozen artifact $\mathcal{A}$. The paired entry-removed condition deletes only $a$ from that context, keeping all other supplied entries unchanged. For retrieval-based systems, both conditions use the same retrieved context, with only the target atomic claim removed in the paired condition. For systems that supply the entire artifact, the paired condition supplies $\mathcal{A}_{-a}=\mathcal{A}\setminus\{a\}$. This measures the effect of retaining $a$ while keeping the rest of the supplied context fixed.

\subsection{Repeated Paired Evaluations and Harmfulness Classification}
\label{sec:appendix-harmful-statistics}

For each fixed artifact $\mathcal{A}$ and each eligible entry $a\in\mathcal{A}$ with $\mathcal{I}(a)\ne\varnothing$, we perform five paired evaluations on the subsequent eligible task instances, indexed by evaluation seed $s\in\{0,1,2,3,4\}$. The same normalized artifact $\mathcal{A}$, target atomic claim $a$, and claim-removed artifact $\mathcal{A}_{-a}$ are used in all five pairs. Within each seed, the two conditions are evaluated on the same eligible task instances with matched task settings and environment action budgets. Thus, each atomic claim is assessed within its own frozen artifact across all five evaluation seeds. For each seed, we first average per-instance scores over the eligible instances in the prescribed evaluation window, using the reward definitions in Appendix~\ref{sec:appendix-scoring}. We then compute the paired score difference
\begin{equation}
    d_s(a)=R_s(\mathcal{A})-R_s(\mathcal{A}_{-a}),
\end{equation}
where $R_s(\mathcal{A})$ and $R_s(\mathcal{A}_{-a})$ are the mean scores on the same entry-specific task set $\mathcal{I}(a)$ using the complete artifact and the artifact with only $a$ removed, respectively. For this fixed artifact and entry, the estimated effect is the mean of the five paired differences, $\bar{d}(a)$.

Following the uncertainty convention in Appendix~\ref{sec:appendix-runs}, we compute a 95\% Student-$t$ confidence interval from the variation in paired score differences across seeds:
\begin{equation}
    \bar{d}(a)\;\pm\;t_{0.975,4}\frac{\hat{\sigma}_d(a)}{\sqrt{5}},
\end{equation}
where $\hat{\sigma}_d(a)$ is the sample standard deviation of the five differences. An entry is classified as harmful only when the entire interval lies below zero.

\paragraph{Harmful-entry rates and aggregation.}
For artifact $\mathcal{A}_j$, let $E_j$ be the number of eligible entries and $H_j$ the number classified as harmful under the five-seed paired evaluation above. Its harmful-entry rate is $h_j=H_j/E_j$. All eligible atomic claims contribute to $E_j$; $H_j$ counts the claims satisfying the harmfulness criterion. For each method and task stream, we report the unweighted arithmetic mean of the three artifact-level rates,
\begin{equation}
    \bar{h}=\frac{1}{3}\sum_{j=1}^{3}h_j.
\end{equation}
Table~\ref{tab:harmful-entry-rate} expresses $\bar{h}$ as a percentage. We average the three rates rather than pooling entries across artifacts; each artifact receives equal weight regardless of its number of eligible entries. The three artifacts and the five evaluation seeds serve distinct roles: harmfulness is assessed for each entry using five paired evaluations within its fixed artifact, then the resulting rates are averaged across artifacts.

\subsection{Reliability of Admission Routing}
\label{sec:appendix-routing-reliability}

To directly assess the reliability contribution of admission routing, we apply the harmful-entry protocol above to RIKA and the three routing variants in Section~\ref{sec:analysis-routing}: Immediate-only admits every claim immediately, Deferred-only defers every claim, and Random Router assigns routes randomly while preserving RIKA's routing ratio. All four use DeepSeek-V4-Flash on \textsc{DB}, \textsc{Cohort}, and \textsc{Codebase}. The evaluated artifact is the frozen knowledge file supplied to the task agent, decomposed into atomic claims using the same normalization as above. The evaluation covers the claims exposed to the task agent.

We use exactly the same atomic-claim normalization, freezing points, subsequent evaluation windows, claim-removal conditions, and harmfulness criterion as in Appendix~\ref{sec:appendix-harmful-evaluation}. For each variant and task, we evaluate three frozen artifacts, with five paired evaluation seeds for every eligible entry in each fixed artifact. An entry is harmful only if the 95\% confidence interval of its paired performance difference lies entirely below zero. Table~\ref{tab:routing-harmful-entry-rate} reports the unweighted mean of the three artifact-level harmful-entry percentages. This comparison measures the harmfulness of retained knowledge and complements the task-performance results in Table~\ref{tab:routing-ablation}.

\begin{table}[!htbp]
\centering
\small
\setlength{\tabcolsep}{5pt}
\begin{tabular}{lccc}
\toprule
\textbf{Method} & \textbf{\textsc{DB}} & \textbf{\textsc{Cohort}} & \textbf{\textsc{Codebase}} \\
\midrule
RIKA & 3.7 & 6.2 & 5.7 \\
Immediate-only & 22.1 & 16.3 & 20.1 \\
Deferred-only & 5.9 & 5.3 & 6.9 \\
Random Router & 14.2 & 19.6 & 15.2 \\
\bottomrule
\end{tabular}
\caption{\textbf{Harmful-entry rates (\%) under admission-routing ablations.} All variants follow the same harmful-entry evaluation protocol.}
\label{tab:routing-harmful-entry-rate}
\end{table}

\section{EKA with Alternative Extractors}
\label{sec:appendix-transfer}

\paragraph{Integration.}
The adapted variants use only the extraction component of SkillOpt or Trace2Skill in place of RIKA's hierarchical extractor. All subsequent stages use EKA: atomization, immediate/deferred routing, semantic relation judgment, evidence accumulation, admission and forgetting, and knowledge-state rendering and context injection. 

\paragraph{Evaluation protocol.}
We follow the setup in Section~\ref{sec:experimental-setup} and Appendix~\ref{sec:appendix-runs}, using DeepSeek-V4-Flash on \textsc{DB}, \textsc{Cohort}, and \textsc{Codebase}. Table~\ref{tab:admission-transfer} reports mean stream-level scores over five runs with the same fixed corpus and run-indexed task ordering as the main experiments. Both adapted variants use RIKA's default downstream settings: batch size $n=5$, immediate threshold $1$, deferred threshold $k=3$, and forgetting window $m=3$. Context management and model settings follow Appendix~\ref{sec:appendix-model-config}.

The comparison with each original system measures the combined effect of replacing its downstream processing with EKA, including atomization and evidence-based reuse. Section~\ref{sec:analysis-routing} examines admission routing within RIKA.

\section{Prompts and Worked Example}
\label{sec:appendix-prompts}

This section specifies RIKA's task-independent prompts, operational instructions, and output fields. Extraction prompts receive only the current trajectory batch and, for second-order extraction, the corresponding first-order outputs. All components are instructed to preserve evidential scope and to avoid filling missing details with model knowledge.

\subsection{Prompt Templates and Output Schemas}
\label{sec:appendix-prompt-templates}

\paragraph{Feedback-grounding requirement.}
All extracted claims must be grounded, directly or indirectly, in environment feedback contained in the source trajectories. Here, environment feedback includes environment observations, tool outputs, errors, and task outcomes. Each candidate must identify the relevant feedback and its connection to the claim; actions and agent reasoning provide context but do not independently justify extraction. Indirect grounding may involve inference or abstraction through factual or procedural claims, but must remain traceable to the original feedback. Claims without such a connection are excluded before admission routing.

\paragraph{Shared extraction instruction.}
Each extractor receives the following common instruction, followed by a type-specific objective:
\begin{quote}\small
Extract only reusable claims that have a direct or indirect grounding in environment feedback in the supplied trajectories. Environment feedback includes environment observations, tool outputs, errors, and task outcomes. Start from this feedback and identify what it establishes or reasonably suggests. Actions, plans, and agent reasoning may provide context, but cannot independently ground a claim; do not extract generic knowledge merely because it appears in the trajectory. For every claim, cite the minimal feedback spans, mark the grounding as direct or indirect, and explain the connection. For indirect claims, state the inference or abstraction and what remains unverified; trace any first-order parent claims back to their original feedback. Write self-contained claims whose scope states the task, entities, conditions, and environment version needed for reuse. Preserve uncertainty and exceptions. Do not infer success from an attempted action or present a hypothesis as an observed fact. Exclude every claim without this feedback connection. Return an empty list if no eligible reusable claim exists.
\end{quote}

The factual and procedural extractors receive a serialized \texttt{trajectories} field. The conceptual extractor additionally receives \texttt{factual\_items}, and the metacognitive extractor additionally receives \texttt{procedural\_items}. These first-order items include their IDs, scopes, feedback spans, grounding explanations, and provenance fields.

The four type-specific objectives are:
\begin{itemize}[leftmargin=*, itemsep=2pt, topsep=4pt]
    \item \textbf{Factual extractor:} identify concrete properties of entities, schemas, states, tool behavior, feedback, and observed outcomes. Cite the environment feedback establishing or suggesting each property, and keep statements tied to the versions and conditions actually observed.
    \item \textbf{Procedural extractor:} identify executable operations or workflows whose behavior is grounded in environment feedback in the trajectories. Specify ordered actions, required inputs or tools, prerequisites, expected outputs, and observed failure-recovery steps. Link the relevant actions to observed tool outputs, errors, or outcomes. Action sequences alone do not qualify. Do not claim that a procedure is generally effective merely because it appeared in a successful trajectory.
    \item \textbf{Conceptual extractor:} given the trajectories and extracted factual claims, identify relations or regularities that organize multiple facts. Each conceptual claim must cite the factual claim IDs on which it depends, trace them to the original feedback, and explain the abstraction connecting that feedback to the claim. It must not merely restate one fact.
    \item \textbf{Metacognitive extractor:} given the trajectories and extracted procedural claims, identify evidence-grounded guidance about when to select, avoid, monitor, or revise a procedure. Each metacognitive claim must cite the procedural claim IDs on which it depends, trace them to the original feedback, and explain how the observed success, failure, or recovery motivates the guidance.
\end{itemize}

Each extracted item follows a shared field-level schema:
\begin{itemize}[leftmargin=*, itemsep=2pt, topsep=4pt]
    \item \texttt{id}, \texttt{type}, and \texttt{claim}: a local identifier, one of the four knowledge types, and a self-contained claim.
    \item \texttt{scope}: the task family, entities, conditions, prerequisites, and environment or schema version under which the claim is stated.
    \item \texttt{source\_trajectories} and \texttt{evidence\_spans}: immutable trajectory IDs and the minimal environment-feedback spans grounding the claim; action or reasoning context may accompany but cannot replace those spans.
    \item \texttt{grounding\_type} and \texttt{grounding\_explanation}: direct or indirect grounding, and a short explanation linking the cited feedback to the exact claim, including any inference, abstraction, or unverified extension.
    \item \texttt{parent\_ids}: empty for first-order factual and procedural claims; factual IDs for conceptual claims; and procedural IDs for metacognitive claims.
\end{itemize}

\paragraph{Atomization.}
The atomizer is conditioned on the knowledge type and receives one extracted claim at a time:
\begin{quote}\small
Decompose the extracted claim into the smallest claims that can each stand alone and be reused independently. Split conjunctions only when each part remains meaningful with its own scope and evidence. Do not split inseparable steps, remove qualifiers, broaden scope, or introduce new content. For each atom, retain only the feedback spans that ground that atom, preserve the relevant parent lineage, and update the direct or indirect grounding explanation. Exclude an atom if no feedback span directly or indirectly grounds it. If the extracted claim is already atomic and feedback-grounded, return it unchanged.
\end{quote}
For each output atomic claim, the schema contains \texttt{atom\_id}, \texttt{parent\_id}, \texttt{type}, \texttt{claim}, \texttt{scope}, \texttt{source\_trajectories}, \texttt{evidence\_spans}, \texttt{grounding\_type}, and \texttt{grounding\_explanation}.

\paragraph{Admission routing.}
For every eligible atomic claim, the router receives the claim, its cited environment-feedback spans, and its grounding explanation:
\begin{quote}\small
First verify that the claim has a direct or indirect connection to the cited environment feedback. Exclude a claim lacking this connection rather than routing it to \textsc{Deferred}. For an eligible claim, decide whether that feedback establishes the exact claim under its stated scope. Route it to \textsc{Immediate} only when environment observations, tool outputs, or explicit feedback are sufficient without extrapolation, assumed causality, or unavailable information. Route it to \textsc{Deferred} only when it is a feedback-grounded inference or generalization whose full scope needs corroboration, such as an extrapolation beyond the observed conditions or a causal interpretation not yet isolated. Knowledge type alone must not determine the route.
\end{quote}
For eligible claims, the router returns \texttt{atom\_id}, \texttt{route}, \texttt{witness\_trajectory\_ids}, \texttt{witness\_spans}, and a short \texttt{justification} identifying either the direct warrant or both the feedback basis and the extension still requiring support.

\paragraph{Relation judgment.}
The relation judge receives a new atomic claim and relevant canonical claims from the persistent state:
\begin{quote}\small
Compare the claims under their explicit scopes. Return every material relation among \textsc{Equivalent}, \textsc{Specialize}, \textsc{Generalize}, \textsc{Contradict}, \textsc{Overlap}, and \textsc{Independent}. Treat two claims as equivalent only if they entail each other under the same scope. For directional entailment, identify which claim is more specific. Mark contradiction only when the scopes overlap and both claims cannot hold there. Shared vocabulary alone is not a relation. Cite the phrases and scope conditions that determine the judgment; do not alter either claim.
\end{quote}
Each judgment contains \texttt{new\_id}, \texttt{stored\_id}, a set \texttt{relations}, \texttt{entailment\_direction} when applicable, \texttt{overlap\_scope}, and \texttt{rationale}. These judgments feed into the evidence-count and admission updates specified in Table~\ref{tab:state-transitions}.

\paragraph{Programmatic rendering and task-agent injection.}
Rendering programmatically concatenates the admitted claims. After each batch, admission is recomputed with thresholds of $1$ for immediate claims and $k$ for deferred claims. The program groups all currently admitted claims under the four knowledge-type headings and serializes each canonical claim once as a scoped bullet, as described in Appendix~\ref{sec:appendix-rendering}. Stored claim text and its scope, prerequisites, exceptions, and environment version are preserved; internal IDs, evidence counts, provenance, routing decisions, and relation labels are omitted. Empty type headings are omitted. The resulting document replaces the previous knowledge block at each batch boundary.

\subsection{End-to-End Worked Example}
\label{sec:appendix-worked-example}

The following stylized \textsc{DB} example illustrates the information flow across three successive batches, each containing one relevant trajectory, with $k=3$.

In the first batch, $\tau_a$ inspects schema version~1 and observes an \texttt{orders(status, amount)} table with \texttt{completed} and \texttt{pending} orders. For a realized-revenue query, summing all matching orders gives an incorrect answer, whereas filtering \texttt{status = completed} gives the correct answer. In the second batch, $\tau_b$ observes the same contrast on different order records under another query condition in version~1. Neither trajectory involves a schema change. After migration, $\tau_c$ attempts the old query, which fails because the old fields are unavailable. Inspection reveals version~2 fields \texttt{state} and \texttt{value}, with labels \texttt{settled} and \texttt{pending}. Summing all matching orders again gives an incorrect answer, whereas filtering \texttt{state = settled} gives the correct answer.

\paragraph{Hierarchical extraction and atomization.}
From $\tau_a$, the factual extractor records the schema fields, lifecycle values, and feedback on the aggregation results. A compound schema description is atomized into separate claims about the lifecycle field and amount field, preserving their version scope and evidence spans. The procedural extractor records the demonstrated operation: in version~1, sum \texttt{amount} over orders satisfying the query conditions and \texttt{status = completed}.

Conditioned on the factual outputs, the conceptual extractor proposes: \emph{For realized-revenue queries in this order-processing task stream, finalized orders contribute to revenue, whereas pending orders do not.} This relates the observed lifecycle states to their revenue contribution. Its content and scope remain unchanged across all three batches and describe the relationship between lifecycle states and revenue contribution.

The metacognitive claim is first extracted from $\tau_c$, conditioned on the procedural record of failed reuse, schema inspection, and successful query revision: \emph{When a previously successful query fails after a schema change, re-inspect the current schema before reusing the earlier query procedure.} The conceptual and metacognitive claims retain links to their factual and procedural parents, respectively. The metacognitive claim is supported by feedback from the schema-change recovery in $\tau_c$.

\paragraph{Routing decisions.}
Version-specific schema facts are routed to \textsc{Immediate} because inspection directly establishes them. The demonstrated filtered-sum procedures are also immediate under the observed task conditions. The conceptual claim is deferred because its task-stream scope extends beyond the records and conditions observed in one trajectory. The metacognitive claim is deferred when it appears in the third batch because it generalizes one recovery into guidance for future schema changes. The schema-inspection output directly grounds the version-specific facts; the query-result feedback directly grounds the procedures under the observed conditions. The same feedback indirectly grounds the conceptual generalization through its factual parents, while the error and recovery feedback indirectly grounds the metacognitive guidance through its procedural parents. These decisions depend on evidence and scope, not knowledge type.

For comparison, suppose the extractor also proposes \emph{All database versions in this task stream use \texttt{\upshape completed} as the finalized label.} Despite being factual in form, this overbroad claim is deferred because a version~1 observation does not establish its cross-version scope.

By contrast, a proposed claim that adding an index makes the query faster is excluded before routing: these trajectories contain no timing or query-plan feedback from which to derive it. An agent's suggestion to add an index would not supply the missing environment feedback. This differs from the fixed-label claim, which has a feedback basis but overgeneralizes its scope.

\paragraph{Subsequent support and counterevidence.}
The conceptual claim is independently re-derived from each trajectory and consolidated into entry $v_C$. Each trajectory supplies one fresh source ID, increasing its evidence count from $1$ to $2$ to $3$. It reaches $k=3$ and is admitted after the third batch. Each trajectory contributes one source-level evidence unit, with parent claims and duplicate derivations linked to that same source.

The metacognitive entry $v_M$ is created only after $\tau_c$, with $e(v_M)=1$, and remains pending. The schema-change recovery in $\tau_c$ supplies the supporting trajectory for this guidance.

In this illustration, the overbroad fixed-label claim is also re-derived from $\tau_b$, raising its count from $1$ to $2$. The observation of \texttt{settled} in $\tau_c$ provides opposing evidence, reducing it to $1$. It remains pending and is not rendered. The version~1 fact remains valid within its historical scope, which does not overlap the version~2 counterexample. New version~2 facts and the demonstrated query procedure are admitted through the immediate route.

\paragraph{Batch-wise reassessment.}
To illustrate the same rule for a previously admitted claim, consider a deferred claim with $e_j(v)=3$ and $k=3$. If the next batch supplies one opposing trajectory and no new support, then $e_{j+1}(v)=2$. The claim remains in the state but is omitted from the next knowledge file. If a later supporting trajectory arrives before the claim is forgotten, it restores the cumulative count to $3$, making the claim eligible again and resetting its below-threshold counter. An immediate claim uses the same rule with threshold $1$: a decrease from $1$ to $0$ makes it pending and starts its below-threshold counter. For either route, the claim is forgotten only after $m$ consecutive batches below its admission threshold. A batch whose updated evidence remains below threshold increments this counter.

\paragraph{Rendered knowledge.}
After the third batch, the admitted entries produce the following document:
\begin{quote}\small
\textbf{Factual Knowledge}
\begin{itemize}[leftmargin=*, itemsep=1pt, topsep=2pt]
    \item In schema version~1, \texttt{orders.status} includes \texttt{completed} and \texttt{pending}.
    \item In schema version~1, order amounts are stored in \texttt{orders.amount}.
    \item In schema version~2, \texttt{orders.state} includes \texttt{settled} and \texttt{pending}.
    \item In schema version~2, order amounts are stored in \texttt{orders.value}.
\end{itemize}
\textbf{Conceptual Knowledge}
\begin{itemize}[leftmargin=*, itemsep=1pt, topsep=2pt]
    \item For realized-revenue queries in this order-processing task stream, finalized orders contribute to revenue, whereas pending orders do not.
\end{itemize}
\textbf{Procedural Knowledge}
\begin{itemize}[leftmargin=*, itemsep=1pt, topsep=2pt]
    \item In schema version~1, sum \texttt{amount} over orders satisfying the query conditions and \texttt{status = completed} for the observed realized-revenue task.
    \item In schema version~2, sum \texttt{value} over orders satisfying the query conditions and \texttt{state = settled} for the corresponding observed task.
\end{itemize}
\end{quote}
The metacognitive and overbroad fixed-label claims remain pending and are omitted, along with internal provenance, evidence counts, and relation annotations. The rendered document includes headings for knowledge types represented in the admitted set. Each entry is assessed using evidence relevant to its own content and scope.

\section{Algorithm and State-Update Details}
\label{sec:appendix-algorithm}

Algorithm~\ref{alg:rika-update} and Table~\ref{tab:state-transitions} specify the updates in Sections~\ref{sec:relation}--\ref{sec:render}. Both admission routes use the same evidence-update procedure, with thresholds of $1$ for immediate claims and $k$ for deferred claims.

\subsection{Claim Schema and Persistent State}
\label{sec:appendix-state-schema}

The state $S_j=(V_j,E_j,H_j,\mathcal{L}_j)$ contains canonical claims, entailment edges, other relation annotations, and an observation ledger. Each stored claim has:
\begin{itemize}[leftmargin=*, itemsep=2pt, topsep=4pt]
    \item \textbf{Content and scope:} canonical ID, atomic content, aliases, knowledge type, task family, entities, prerequisites, and environment/schema version.
    \item \textbf{Provenance and evidence:} source trajectory IDs, batches, environment-feedback spans, direct or indirect grounding and its explanation, extraction lineage tracing parent claims to original feedback, batch-level support and opposition counts, and cumulative net evidence $e(v)$.
    \item \textbf{Routing and activity:} admission route $z(v)$ and its source justification; creation and last-evidence batches; and below-threshold counter. Admission is derived from the current evidence count and route after each batch.
    \item \textbf{Relations:} incident edges and judgments with their provenance.
\end{itemize}
The admitted set is recomputed from the claims in $V_j$. Claims below their admission threshold remain pending until they meet the forgetting condition. Forgetting removes the claim, its stored evidence, and its incident relations from the persistent state. If equivalent knowledge is extracted again later, it is processed as a new claim.

\subsection{Batch Evidence Counts and Graph Updates}
\label{sec:appendix-evidence-ledger}

After each batch, the model outputs evidence counts based on the batch trajectories and the semantic relations between extracted and stored claims. A trajectory contributes support when new environment observations or task feedback ground an equivalent claim or a claim that entails the stored claim. Extraction is conditioned on the current trajectory batch and the corresponding first-order outputs. Evidence increments require new environment observations or task feedback supporting the claim.

Same-scope equivalent claims are consolidated, with their provenance and graph relations retained. If their admission routes disagree, the merged claim falls back to \textsc{Deferred}. In particular, an existing deferred claim remains deferred when a newly extracted equivalent claim is immediate. Changed scopes remain separate. Entailment edges point from more specific to more general claims, and the graph remains acyclic.

Let $c_j^+(v)$ and $c_j^-(v)$ denote the supporting and opposing evidence counts assigned from batch $j$. The cumulative net evidence is updated as
\begin{equation}
e_j(v)=e_{j-1}(v)+c_j^+(v)-c_j^-(v),
\label{eq:batch-evidence}
\end{equation}
with initial count zero for a new claim before its source evidence is incorporated. The main text's $\pm1$ rules describe a single supporting or opposing observation. Support follows entailment from specific claims to reachable generalizations; opposition requires a grounded contradiction within the target's scope. All active claims, including previously admitted claims, receive evidence updates. Fresh evidence is grounded in the current batch trajectories.

\begin{table*}[t]
\centering
\small
\setlength{\tabcolsep}{4pt}
\begin{tabular}{p{0.15\textwidth} p{0.29\textwidth} p{0.50\textwidth}}
\toprule
\textbf{Relation or event} & \textbf{Claim and graph update} & \textbf{Evidence and admission update} \\
\midrule
\textsc{Equivalent} & Merge claims and rewire edges. & Incorporate batch support counts; conflicting routes fall back to deferred. \\
\textsc{Specialize} & Keep both; add $u\rightarrow v$. & Transfer support along $u\rightarrow v$ and onward to reachable generalizations. \\
\textsc{Generalize} & Keep both; add $v\rightarrow u$. & Transfer support along $v\rightarrow u$ and onward to reachable generalizations. \\
\textsc{Contradict} & Keep both; annotate conflict in $H_j$. & Apply the batch opposition count to the target; assess both claims by their own thresholds. \\
\textsc{Overlap} & Keep both; annotate overlap in $H_j$. & Preserve each claim's evidence count. \\
\textsc{Independent} & Keep claims separate and preserve their existing relations. & Initialize unmatched claims from their batch evidence counts; leave existing evidence unchanged. \\
\midrule
Admission & Recompute the admitted set after every batch. & A surviving claim is admitted iff $e(v)\geq\theta(v)$, where $\theta(v)=1$ for immediate and $\theta(v)=k$ for deferred claims. \\
Pending & Retain a surviving claim below its threshold. & Preserve accumulated evidence and route across pending batches. \\
Forgetting & Delete the claim and its incident relations. & Forget after $m$ consecutive batches with $e_j(v)<\theta(v)$. Reaching the threshold resets the counter; a below-threshold batch increments it. \\
\bottomrule
\end{tabular}
\caption{State updates for a new claim $u$ and stored claim $v$. Evidence updates precede forgetting and batch-wise admission.}
\label{tab:state-transitions}
\end{table*}

\subsection{Batch-wise Admission and Forgetting}
\label{sec:appendix-lifecycle}

For a stored claim $v$, the route-specific threshold is
\begin{equation}
\theta(v)=
\begin{cases}
1, & z(v)=\textsc{Immediate},\\
k, & z(v)=\textsc{Deferred}.
\end{cases}
\label{eq:route-threshold}
\end{equation}
After updating evidence and applying forgetting, recompute the admitted set
\begin{equation}
A_j=\{v\in V_j:e_j(v)\geq\theta(v)\}.
\label{eq:batch-admission}
\end{equation}
Here $e_j(v)$ is the current cumulative net evidence after batch $j$. Immediate claims therefore need one net unit of evidence, while deferred claims need at least $k$. The route is retained with the claim, with disagreements during equivalent-claim merging resolved as deferred; admission is recalculated from the updated state after every batch. Previously admitted claims receive the same support and opposition updates as other active claims. Counterevidence reduces the cumulative net evidence used in the batch-wise admission decision. A surviving claim can be included again whenever its cumulative evidence meets its threshold.

A claim is forgotten if its evidence count remains below its admission threshold for $m$ consecutive batches. Let $d_j(v)$ denote its number of consecutive below-threshold batches. After incorporating batch $j$:
\begin{equation}
d_j(v)=\begin{cases}
d_{j-1}(v)+1, & e_j(v)<\theta(v),\\
0, & e_j(v)\geq\theta(v).
\end{cases}
\label{eq:forgetting-counter}
\end{equation}
The counter is initialized to zero when a claim is created, before evaluating its first batch. A claim first below threshold in batch $j$ is therefore forgotten after batch $j+m-1$ if it remains below threshold throughout. Each batch whose updated evidence count remains below threshold increments the counter, as specified in Equation~\ref{eq:forgetting-counter}. Until the counter reaches $m$, the claim remains pending and is excluded from the rendered file. At $d_j(v)\geq m$, delete the claim, its evidence, and its incident relations. Later extraction of equivalent knowledge initializes a new claim and its evidence and activity counters.

\paragraph{Configuration.}
The default batch size is $n=5$, the deferred admission threshold is $k=3$, the immediate threshold is $1$, and the forgetting window is $m=3$ completed batches. Thus, a claim is forgotten after three consecutive batches below its admission threshold; reaching the threshold resets the counter.

\subsection{End-to-End Pseudocode}
\label{sec:appendix-pseudocode}

Algorithm~\ref{alg:rika-update} combines model-based extraction, atomization, routing, and relation judgment with model-produced batch evidence counts, cumulative state updates, and batch-wise admission.

\begin{algorithm*}[t]
\small
\caption{Batched RIKA updates}
\label{alg:rika-update}
\begin{algorithmic}[1]
\Require Ordered stream $\mathcal{T}$; batch size $n$; deferred threshold $k$; forgetting window $m$
\State $S_0\gets(\emptyset,\emptyset,\emptyset,\emptyset)$; $M_0\gets\text{empty file}$
\For{each consecutive batch $j$, including a final partial batch}
    \State $B_j\gets\Call{SolveTasks}{\mathcal{T},n,M_{j-1}}$; assign immutable trajectory IDs and observation batches
    \State $F_j\gets\mathcal{E}_F(B_j)$; $P_j\gets\mathcal{E}_P(B_j)$; $C_j\gets\mathcal{E}_C(B_j,F_j)$; $G_j\gets\mathcal{E}_M(B_j,P_j)$
    \State $U_j\gets\Call{AtomizeWithProvenance}{F_j\cup P_j\cup C_j\cup G_j}$
    \State Retain in $U_j$ only atoms with direct or indirect environment-feedback grounding
    \State $Z_j\gets\Call{Route}{U_j,B_j}$
    \State $J_j\gets\Call{JudgeRelations}{U_j,V_{j-1}}$, including new--new pairs
    \State $S\gets\Call{MergeEquivalents}{S_{j-1},U_j,J_j}$
    \State $S\gets\Call{ApplySixRelations}{S,J_j,Z_j}$; resolve route disagreements as deferred and check cycles
    \State $(c_j^+,c_j^-)\gets\Call{ModelEvidenceCounts}{B_j,U_j,J_j,S}$
    \State Update $e_j(v)\gets e_{j-1}(v)+c_j^+(v)-c_j^-(v)$ for all affected claims
    \For{each active claim $v$}
        \State $\theta(v)\gets 1$ if $z(v)=\textsc{Immediate}$; otherwise $\theta(v)\gets k$
        \If{$e_j(v)<\theta(v)$}
            \State $d_j(v)\gets d_{j-1}(v)+1$ \Comment{New claims start with counter zero}
        \Else
            \State $d_j(v)\gets0$
        \EndIf
        \If{$d_j(v)\geq m$}
            \State Delete $v$, its evidence, and its incident relations from $S$
        \EndIf
    \EndFor
    \State $S_j\gets S$
    \State $A_j\gets\{v\in V_j:e_j(v)\geq\theta(v)\}$ \Comment{Recompute from current state}
    \State $M_j\gets\Call{SerializeByType}{A_j}$; replace the knowledge block for the next batch
\EndFor
\State \Return final $S_j,M_j$
\end{algorithmic}
\end{algorithm*}

Atomic claims enter routing after their direct or indirect connection to environment feedback is established. Each surviving atom preserves the relevant feedback spans and grounding explanation; second-order claims remain traceable through their parents to the original feedback. Unresolved relations are re-judged or withheld. Extraction uses the current trajectories and corresponding first-order outputs. Each claim's admission route is determined by its cited environment feedback, while entailment edges transfer support. Forgetting and serialization follow Appendices~\ref{sec:appendix-lifecycle} and~\ref{sec:appendix-rendering}.

\subsection{Deterministic Rendering and Context Injection}
\label{sec:appendix-rendering}

The renderer programmatically concatenates all claims in the current admitted set $A_j$, grouped by knowledge type. Each canonical claim is included once with its scope and qualifications; internal evidence records and routing metadata are omitted. An empty admitted set yields an empty file.

Insert the complete $M_j$ after system/tool instructions and before each task in $B_{j+1}$ as scoped guidance, replacing the previous knowledge block. Keep it fixed within the batch and preserve it during history compaction. Completed trajectories update the next file, and the complete admitted knowledge document is supplied to every task in the following batch.

\section{Experimental Setup and Reproducibility}
\label{sec:appendix-setup}

All experiments on Continual Learning Bench strictly follow the benchmark protocols and settings provided in its official GitHub repository~\citep{asawa2026continual}, including task-stream construction, stage boundaries, within-stage permutations, environment action budgets, and reward computation. We apply these benchmark-defined protocols consistently across RIKA and all baselines. The following subsections summarize the benchmark settings and separately describe RIKA's model configuration and adaptation procedure.

\subsection{Task Streams}
\label{sec:appendix-streams}

We evaluate continual adaptation on six ordered task streams from Continual Learning Bench~\citep{asawa2026continual}:
\begin{itemize}[leftmargin=*, itemsep=2pt, topsep=4pt]
    \item \textbf{Blind Spectrum Monitoring (\textsc{BSM})} contains 90 scans divided into three fixed stages of 30 scans each. The stream progresses from five predominantly active wideband channels to a mixed wideband/narrowband grid, and finally to the full 13-channel grid.
    \item \textbf{Cohort Studies (\textsc{Cohort})} contains 20 databases from five studies in the fixed order HERALD, MERIDIAN, MOSAIC, FORGE, and CADENCE, with four region/draw instances per study.
    \item \textbf{Database Exploration (\textsc{DB})} contains 40 questions: 20 before and 20 after a schema migration.
    \item \textbf{Exploitable Poker (\textsc{Poker})} contains 120 hands across five opponent stages of lengths 20, 30, 10, 35, and 25.
    \item \textbf{Sales Prediction (\textsc{Sales})} contains 12 chronological forecasting instances covering 2027--2038.
    \item \textbf{Codebase Adaptation (\textsc{Codebase})} contains nine Tablib issues followed by ten Tenacity issues.
\end{itemize}
Stage order and boundaries remain fixed across runs. Repeated runs permute instances only within stages, except for \textsc{Sales}, whose chronological order is preserved in every run.

\subsection{Scoring}
\label{sec:appendix-scoring}

For each task stream, the run-level score is the arithmetic mean of its per-instance rewards. Rewards are defined as follows:
\begin{itemize}[leftmargin=*, itemsep=2pt, topsep=4pt]
    \item \textbf{\textsc{BSM}:} interval-set intersection over union (IoU) over long-run available spectrum.
    \item \textbf{\textsc{Cohort}:} the reduction in mean binary survival Kullback--Leibler (KL) divergence, measured in bits, relative to a study-wide flat predictor.
    \item \textbf{\textsc{DB}:} $1-q/15$ for a correct answer submitted after $q$ exploratory queries, and zero for an incorrect answer or an exhausted query budget.
    \item \textbf{\textsc{Poker}:} chip profit divided by the big blind.
    \item \textbf{\textsc{Sales}:} unclipped WAPE skill, computed as $1-\frac{\sum_i |\hat{y}_i-y_i|}{\sum_i |y_i|}$, where WAPE denotes weighted absolute percentage error and $\hat{y}_i$ and $y_i$ are the predicted and observed values, respectively.
    \item \textbf{\textsc{Codebase}:} $1-(t-1)/40$ for an issue solved at command $t$, and zero for an unsolved issue.
\end{itemize}
Because rewards have heterogeneous units and ranges, we normalize each task before averaging across streams, following Continual Learning Bench~\citep{asawa2026continual}. For a fixed backbone and method in Table~\ref{tab:main-results}, let $R_{t,s}$ be the mean per-instance reward of learning run $s$ on task $t$, and let $c_t$ be the shared stateless score for that backbone and task, averaged over the five ICL stateless control runs (Table~\ref{tab:paired-stateless}). The same $c_t$ is used for all methods under that backbone. Let $U_t$ be the benchmark-defined maximum reward and $b_t^{\mathrm{ext}}$ the fixed external reference score, taken from the GPT-5.4 stateless baseline in the benchmark's ICL reference run. This external reference is shared across backbones for score normalization, whereas $c_t$ is backbone-specific and is used to measure gain.

With $T=6$ task streams and $S=5$ runs per task, we report
\begin{align}
\operatorname{AggScore}
&=\frac{1}{TS}\sum_{t=1}^{T}\sum_{s=0}^{S-1}
\frac{R_{t,s}-b_t^{\mathrm{ext}}}{U_t-b_t^{\mathrm{ext}}},
\label{eq:aggregate-score}\\
\operatorname{AggGain}
&=\frac{1}{TS}\sum_{t=1}^{T}\sum_{s=0}^{S-1}
\frac{R_{t,s}-c_t}{U_t-c_t}.
\label{eq:aggregate-gain}
\end{align}
The shared control $c_t$ and its remaining headroom $U_t-c_t$ are fixed across methods and learning runs for each backbone and task. We average over the five learning runs and equally over the six tasks. Aggregate score measures performance relative to the fixed external reference, whereas aggregate gain measures improvement relative to the backbone-specific shared stateless control, normalized by its remaining headroom. The two aggregate metrics use their respective denominators in Equations~\ref{eq:aggregate-score} and~\ref{eq:aggregate-gain}. Task-level \textsc{Score} and \textsc{Gain} retain their raw reward units.

\subsection{Model Configuration and Adaptation}
\label{sec:appendix-model-config}
\label{sec:appendix-gemini}

We use \texttt{gemini-3-flash-preview}, the 0731 version of DeepSeek-V4-Flash, MiniMax-M3, and GPT-5.4. Within each experiment, the task agent and RIKA's auxiliary components use the same backbone. All four backbones share the following settings: temperature zero, a maximum output length of 8,192 tokens per call, and a task-agent context limit of 128k tokens with a 500-token reserve.

The task-agent context is cleared between instances, and the current knowledge document is re-injected at the start of each instance. RIKA updates its knowledge state every five completed instances by default; for \textsc{Sales}, which contains only 12 instances, we use $n=4$. Auxiliary calls are recorded separately from the task agent's environment-action budget.

\subsection{Repeated Runs and Uncertainty Estimates}
\label{sec:appendix-runs}

We conduct five runs indexed by $s\in\{0,1,2,3,4\}$. These indices determine reproducible within-stage permutations of the same fixed instance corpus and environments. For \textsc{Sales}, all five runs use the same chronological stream.

We report the mean and sample standard deviation of the five run-level scores. We compute 95\% Student-$t$ confidence intervals with four degrees of freedom as
\begin{equation}
    \bar{R} \;\pm\; t_{0.975,4}\frac{\hat{\sigma}_R}{\sqrt{5}},
\end{equation}
where $\bar{R}$ is the mean run-level score, $\hat{\sigma}_R$ is its sample standard deviation across runs, and $t_{0.975,4}$ is the 97.5th percentile of the Student-$t$ distribution with four degrees of freedom.

\subsection{Shared Stateless Controls}
\label{sec:appendix-stateless}

For the main comparison in Table~\ref{tab:main-results} and the ablation and sensitivity analyses in Appendix~\ref{sec:appendix-additional-analysis}, we use one shared stateless reference for each backbone and task stream. The reference is the mean of the five stateless runs from the corresponding ICL evaluation, using the common no-memory configuration without retaining experience across instances. These control runs are executed independently of the learning methods; their mean is reused for all methods under the same backbone and task.

For a fixed backbone and task stream, let $R_s(m)$ denote the score of method $m$ in run $s$, and let $R_s^{0}$ denote the score of the shared stateless control in run $s$. Define $c=\frac{1}{5}\sum_{s=0}^{4}R_s^{0}$. The reported quantities are
\begin{align}
\operatorname{Score}(m) &= \frac{1}{5}\sum_{s=0}^{4}R_s(m), \\
\operatorname{Gain}(m) &= \operatorname{Score}(m)-c.
\end{align}
Thus, every method is evaluated against the same stateless reference within each backbone--task combination. Table~\ref{tab:paired-stateless} reports these shared reference scores.

For learning dynamics, the cumulative gain over the stateless control is defined as
\[
G_i^{(s)}
=
\sum_{\ell=1}^{i}
\left(
r_{s,\ell}-r^{0}_{s,\ell}
\right),
\]
where \(r_{s,\ell}\) and \(r^{0}_{s,\ell}\) are the rewards of a method and the stateless control at instance \(\ell\) in run \(s\).

\begin{table}[!htbp]
\centering
\small
\setlength{\tabcolsep}{3pt}
\resizebox{\columnwidth}{!}{%
\begin{tabular}{l*{6}{r}}
\toprule
\textbf{Backbone} & \textbf{DB} & \textbf{Poker} & \textbf{Cohort} & \textbf{BSM} & \textbf{Sales} & \textbf{Codebase} \\
\midrule
GPT-5.4 & 0.138 & 1.113 & -0.007 & 0.219 & 0.466 & 0.498 \\
DeepSeek-V4-Flash & 0.060 & 0.652 & -0.118 & 0.219 & 0.514 & 0.512 \\
MiniMax-M3 & 0.136 & -0.135 & -0.005 & 0.220 & 0.197 & 0.373 \\
Gemini-3-Flash & 0.089 & 1.640 & -0.059 & 0.219 & 0.439 & 0.407 \\
\bottomrule
\end{tabular}%
}
\caption{\textbf{Shared stateless performance averaged over five runs.} For each backbone and task stream, the reference is the mean of the five stateless controls from the corresponding ICL evaluation. All methods in Table~\ref{tab:main-results} use this same reference to compute Gain and normalized aggregate Gain; the ablation and sensitivity analyses use the corresponding DeepSeek-V4-Flash reference for Gain.}
\label{tab:paired-stateless}
\end{table}

\section{Main Results with 95\% Confidence Intervals}
\label{sec:appendix-main-ci}

Tables~\ref{tab:main-ci-gpt}--\ref{tab:main-ci-gemini} report the Score and Gain means in Table~\ref{tab:main-results}. Entries with intervals use the format mean $\pm$ 95\% confidence-interval half-width. Score intervals follow the five-run Student-$t$ definition in Appendix~\ref{sec:appendix-runs}; aggregate Score intervals use run-level aggregate scores. Bold means follow the main table and do not indicate statistical significance.

\paragraph{Conditional intervals for task-level Gain.}
We treat each shared stateless reference $c_t$ in Table~\ref{tab:paired-stateless} as fixed. Subtracting this constant shifts the Score interval without changing its half-width: if the Score interval is $[L,U]$, the corresponding Gain interval is $[L-c_t,U-c_t]$. These intervals quantify variation in learning performance conditional on the reported reference; they exclude uncertainty in estimating the stateless reference itself. Aggregate Gain intervals are reported for all four backbones.

\begin{table*}[!tp]
\centering
\scriptsize
\setlength{\tabcolsep}{2.4pt}
\renewcommand{\arraystretch}{1.16}
\resizebox{\textwidth}{!}{%
\begin{tabular}{ll*{7}{c}}
\toprule
\textbf{Method} & \textbf{Metric} & \textbf{DB} & \textbf{Poker} & \textbf{Cohort} & \textbf{BSM} & \textbf{Sales} & \textbf{Codebase} & \textbf{Agg.} \\
\midrule
\textbf{RIKA} & Score & $0.351\,\pm\,0.019$ & $\mathbf{1.247}\,\pm\,0.097$ & $-0.020\,\pm\,0.007$ & $\mathbf{0.579}\,\pm\,0.035$ & $\mathbf{0.808}\,\pm\,0.023$ & $0.551\,\pm\,0.018$ & $\mathbf{0.232}\,\pm\,0.029$ \\
 & Gain & $+0.213\,\pm\,0.019$ & $\mathbf{+0.134}\,\pm\,0.097$ & $-0.013\,\pm\,0.007$ & $\mathbf{+0.360}\,\pm\,0.035$ & $\mathbf{+0.342}\,\pm\,0.023$ & $+0.053\,\pm\,0.018$ & $\mathbf{+0.232}\,\pm\,0.029$ \\
\addlinespace[2pt]
ICL & Score & $0.347\,\pm\,0.018$ & $0.798\,\pm\,0.152$ & $-0.028\,\pm\,0.006$ & $0.513\,\pm\,0.023$ & $0.769\,\pm\,0.014$ & $0.534\,\pm\,0.009$ & $0.183\,\pm\,0.021$ \\
 & Gain & $+0.209\,\pm\,0.018$ & $-0.315\,\pm\,0.152$ & $-0.021\,\pm\,0.006$ & $+0.294\,\pm\,0.023$ & $+0.303\,\pm\,0.014$ & $+0.036\,\pm\,0.009$ & $+0.183\,\pm\,0.021$ \\
\addlinespace[2pt]
Mem0 & Score & $\mathbf{0.431}\,\pm\,0.022$ & $0.612\,\pm\,0.044$ & $-0.034\,\pm\,0.006$ & $0.375\,\pm\,0.034$ & $0.651\,\pm\,0.033$ & $0.584\,\pm\,0.025$ & $0.140\,\pm\,0.027$ \\
 & Gain & $\mathbf{+0.293}\,\pm\,0.022$ & $-0.501\,\pm\,0.044$ & $-0.027\,\pm\,0.006$ & $+0.156\,\pm\,0.034$ & $+0.185\,\pm\,0.033$ & $+0.086\,\pm\,0.025$ & $+0.140\,\pm\,0.027$ \\
\addlinespace[2pt]
ACE & Score & $0.196\,\pm\,0.011$ & $1.196\,\pm\,0.073$ & $\mathbf{-0.004}\,\pm\,0.009$ & $0.220\,\pm\,0.071$ & $0.510\,\pm\,0.043$ & $\mathbf{0.609}\,\pm\,0.028$ & $0.067\,\pm\,0.031$ \\
 & Gain & $+0.058\,\pm\,0.011$ & $+0.083\,\pm\,0.073$ & $\mathbf{+0.003}\,\pm\,0.009$ & $+0.001\,\pm\,0.071$ & $+0.044\,\pm\,0.043$ & $\mathbf{+0.111}\,\pm\,0.028$ & $+0.067\,\pm\,0.031$ \\
\addlinespace[2pt]
Trace2Skill & Score & $0.138\,\pm\,0.018$ & $1.115\,\pm\,0.088$ & $-0.032\,\pm\,0.010$ & $0.318\,\pm\,0.041$ & $0.689\,\pm\,0.017$ & $0.513\,\pm\,0.015$ & $0.071\,\pm\,0.034$ \\
 & Gain & $+0.000\,\pm\,0.018$ & $+0.002\,\pm\,0.088$ & $-0.025\,\pm\,0.010$ & $+0.099\,\pm\,0.041$ & $+0.223\,\pm\,0.017$ & $+0.015\,\pm\,0.015$ & $+0.071\,\pm\,0.034$ \\
\addlinespace[2pt]
SkillOpt & Score & $0.189\,\pm\,0.013$ & $0.971\,\pm\,0.068$ & $-0.024\,\pm\,0.015$ & $0.443\,\pm\,0.055$ & $0.566\,\pm\,0.054$ & $0.607\,\pm\,0.022$ & $0.105\,\pm\,0.030$ \\
 & Gain & $+0.051\,\pm\,0.013$ & $-0.142\,\pm\,0.068$ & $-0.017\,\pm\,0.015$ & $+0.224\,\pm\,0.055$ & $+0.100\,\pm\,0.054$ & $+0.109\,\pm\,0.022$ & $+0.105\,\pm\,0.030$ \\
\bottomrule
\end{tabular}%
}
\caption{\textbf{GPT-5.4 results with 95\% confidence intervals.} Means match Table~\ref{tab:main-results}. Task-level Gain intervals condition on the fixed shared control.}
\label{tab:main-ci-gpt}
\end{table*}

\begin{table*}[!tp]
\centering
\scriptsize
\setlength{\tabcolsep}{2.4pt}
\renewcommand{\arraystretch}{1.16}
\resizebox{\textwidth}{!}{%
\begin{tabular}{ll*{7}{c}}
\toprule
\textbf{Method} & \textbf{Metric} & \textbf{DB} & \textbf{Poker} & \textbf{Cohort} & \textbf{BSM} & \textbf{Sales} & \textbf{Codebase} & \textbf{Agg.} \\
\midrule
\textbf{RIKA} & Score & $\mathbf{0.281}\,\pm\,0.016$ & $\mathbf{1.141}\,\pm\,0.121$ & $-0.042\,\pm\,0.016$ & $0.234\,\pm\,0.039$ & $\mathbf{0.847}\,\pm\,0.039$ & $\mathbf{0.575}\,\pm\,0.020$ & $\mathbf{0.141}\,\pm\,0.031$ \\
 & Gain & $\mathbf{+0.221}\,\pm\,0.016$ & $\mathbf{+0.489}\,\pm\,0.121$ & $+0.076\,\pm\,0.016$ & $+0.015\,\pm\,0.039$ & $\mathbf{+0.333}\,\pm\,0.039$ & $\mathbf{+0.063}\,\pm\,0.020$ & $\mathbf{+0.233}\,\pm\,0.031$ \\
\addlinespace[2pt]
ICL & Score & $0.256\,\pm\,0.010$ & $0.755\,\pm\,0.098$ & $-0.084\,\pm\,0.012$ & $\mathbf{0.518}\,\pm\,0.033$ & $0.639\,\pm\,0.042$ & $0.548\,\pm\,0.016$ & $0.074\,\pm\,0.022$ \\
 & Gain & $+0.196\,\pm\,0.010$ & $+0.103\,\pm\,0.098$ & $+0.034\,\pm\,0.012$ & $\mathbf{+0.299}\,\pm\,0.033$ & $+0.125\,\pm\,0.042$ & $+0.036\,\pm\,0.016$ & $+0.176\,\pm\,0.022$ \\
\addlinespace[2pt]
Mem0 & Score & $0.274\,\pm\,0.022$ & $0.732\,\pm\,0.133$ & $\mathbf{-0.038}\,\pm\,0.017$ & $0.348\,\pm\,0.057$ & $0.522\,\pm\,0.051$ & $0.547\,\pm\,0.014$ & $0.049\,\pm\,0.033$ \\
 & Gain & $+0.214\,\pm\,0.022$ & $+0.080\,\pm\,0.133$ & $\mathbf{+0.080}\,\pm\,0.017$ & $+0.129\,\pm\,0.057$ & $+0.008\,\pm\,0.051$ & $+0.035\,\pm\,0.014$ & $+0.129\,\pm\,0.033$ \\
\addlinespace[2pt]
ACE & Score & $0.135\,\pm\,0.023$ & $0.637\,\pm\,0.114$ & $-0.059\,\pm\,0.016$ & $0.222\,\pm\,0.034$ & $0.563\,\pm\,0.080$ & $0.529\,\pm\,0.021$ & $-0.020\,\pm\,0.034$ \\
 & Gain & $+0.075\,\pm\,0.023$ & $-0.015\,\pm\,0.114$ & $+0.059\,\pm\,0.016$ & $+0.003\,\pm\,0.034$ & $+0.049\,\pm\,0.080$ & $+0.017\,\pm\,0.021$ & $+0.071\,\pm\,0.034$ \\
\addlinespace[2pt]
Trace2Skill & Score & $0.158\,\pm\,0.029$ & $0.858\,\pm\,0.122$ & $-0.053\,\pm\,0.021$ & $0.293\,\pm\,0.045$ & $0.602\,\pm\,0.068$ & $0.538\,\pm\,0.015$ & $0.025\,\pm\,0.041$ \\
 & Gain & $+0.098\,\pm\,0.029$ & $+0.206\,\pm\,0.122$ & $+0.065\,\pm\,0.021$ & $+0.074\,\pm\,0.045$ & $+0.088\,\pm\,0.068$ & $+0.026\,\pm\,0.015$ & $+0.115\,\pm\,0.041$ \\
\addlinespace[2pt]
SkillOpt & Score & $0.147\,\pm\,0.038$ & $0.435\,\pm\,0.104$ & $-0.062\,\pm\,0.019$ & $0.218\,\pm\,0.068$ & $0.658\,\pm\,0.044$ & $0.567\,\pm\,0.018$ & $0.017\,\pm\,0.039$ \\
 & Gain & $+0.087\,\pm\,0.038$ & $-0.217\,\pm\,0.104$ & $+0.056\,\pm\,0.019$ & $-0.001\,\pm\,0.068$ & $+0.144\,\pm\,0.044$ & $+0.055\,\pm\,0.018$ & $+0.113\,\pm\,0.039$ \\
\bottomrule
\end{tabular}%
}
\caption{\textbf{DeepSeek-V4-Flash results with 95\% confidence intervals.} Means match Table~\ref{tab:main-results}. Task-level Gain intervals condition on the fixed shared control.}
\label{tab:main-ci-deepseek}
\end{table*}

\begin{table*}[!tp]
\centering
\scriptsize
\setlength{\tabcolsep}{2.4pt}
\renewcommand{\arraystretch}{1.16}
\resizebox{\textwidth}{!}{%
\begin{tabular}{ll*{7}{c}}
\toprule
\textbf{Method} & \textbf{Metric} & \textbf{DB} & \textbf{Poker} & \textbf{Cohort} & \textbf{BSM} & \textbf{Sales} & \textbf{Codebase} & \textbf{Agg.} \\
\midrule
\textbf{RIKA} & Score & $\mathbf{0.327}\,\pm\,0.013$ & $\mathbf{0.623}\,\pm\,0.099$ & $\mathbf{0.012}\,\pm\,0.009$ & $0.257\,\pm\,0.042$ & $0.758\,\pm\,0.079$ & $\mathbf{0.554}\,\pm\,0.012$ & $\mathbf{0.163}\,\pm\,0.015$ \\
 & Gain & $\mathbf{+0.191}\,\pm\,0.013$ & $\mathbf{+0.758}\,\pm\,0.099$ & $\mathbf{+0.017}\,\pm\,0.009$ & $+0.037\,\pm\,0.042$ & $+0.561\,\pm\,0.079$ & $\mathbf{+0.181}\,\pm\,0.012$ & $\mathbf{+0.239}\,\pm\,0.015$ \\
\addlinespace[2pt]
ICL & Score & $0.284\,\pm\,0.019$ & $0.407\,\pm\,0.103$ & $-0.020\,\pm\,0.007$ & $\mathbf{0.337}\,\pm\,0.038$ & $0.535\,\pm\,0.055$ & $0.483\,\pm\,0.026$ & $0.043\,\pm\,0.012$ \\
 & Gain & $+0.148\,\pm\,0.019$ & $+0.542\,\pm\,0.103$ & $-0.015\,\pm\,0.007$ & $\mathbf{+0.117}\,\pm\,0.038$ & $+0.338\,\pm\,0.055$ & $+0.110\,\pm\,0.026$ & $+0.147\,\pm\,0.012$ \\
\addlinespace[2pt]
Mem0 & Score & $0.303\,\pm\,0.010$ & $0.235\,\pm\,0.177$ & $0.003\,\pm\,0.005$ & $0.289\,\pm\,0.058$ & $0.501\,\pm\,0.056$ & $0.402\,\pm\,0.027$ & $0.018\,\pm\,0.014$ \\
 & Gain & $+0.167\,\pm\,0.010$ & $+0.370\,\pm\,0.177$ & $+0.008\,\pm\,0.005$ & $+0.069\,\pm\,0.058$ & $+0.304\,\pm\,0.056$ & $+0.029\,\pm\,0.027$ & $+0.132\,\pm\,0.014$ \\
\addlinespace[2pt]
ACE & Score & $0.189\,\pm\,0.020$ & $0.509\,\pm\,0.088$ & $-0.026\,\pm\,0.009$ & $0.230\,\pm\,0.051$ & $\mathbf{0.772}\,\pm\,0.072$ & $0.513\,\pm\,0.016$ & $0.082\,\pm\,0.019$ \\
 & Gain & $+0.053\,\pm\,0.020$ & $+0.644\,\pm\,0.088$ & $-0.021\,\pm\,0.009$ & $+0.010\,\pm\,0.051$ & $\mathbf{+0.575}\,\pm\,0.072$ & $+0.140\,\pm\,0.016$ & $+0.159\,\pm\,0.019$ \\
\addlinespace[2pt]
Trace2Skill & Score & $0.175\,\pm\,0.015$ & $0.612\,\pm\,0.142$ & $-0.031\,\pm\,0.007$ & $0.219\,\pm\,0.034$ & $0.638\,\pm\,0.078$ & $0.533\,\pm\,0.027$ & $0.039\,\pm\,0.023$ \\
 & Gain & $+0.039\,\pm\,0.015$ & $+0.747\,\pm\,0.142$ & $-0.026\,\pm\,0.007$ & $-0.001\,\pm\,0.034$ & $+0.441\,\pm\,0.078$ & $+0.160\,\pm\,0.027$ & $+0.128\,\pm\,0.023$ \\
\addlinespace[2pt]
SkillOpt & Score & $0.180\,\pm\,0.024$ & $0.559\,\pm\,0.170$ & $-0.028\,\pm\,0.012$ & $0.234\,\pm\,0.042$ & $0.672\,\pm\,0.080$ & $0.517\,\pm\,0.018$ & $0.050\,\pm\,0.027$ \\
 & Gain & $+0.044\,\pm\,0.024$ & $+0.694\,\pm\,0.170$ & $-0.023\,\pm\,0.012$ & $+0.014\,\pm\,0.042$ & $+0.475\,\pm\,0.080$ & $+0.144\,\pm\,0.018$ & $+0.137\,\pm\,0.027$ \\
\bottomrule
\end{tabular}%
}
\caption{\textbf{MiniMax-M3 results with 95\% confidence intervals.} Means match Table~\ref{tab:main-results}. Task-level Gain intervals condition on the fixed shared control.}
\label{tab:main-ci-minimax}
\end{table*}

\begin{table*}[!tp]
\centering
\scriptsize
\setlength{\tabcolsep}{2.4pt}
\renewcommand{\arraystretch}{1.16}
\resizebox{\textwidth}{!}{%
\begin{tabular}{ll*{7}{c}}
\toprule
\textbf{Method} & \textbf{Metric} & \textbf{DB} & \textbf{Poker} & \textbf{Cohort} & \textbf{BSM} & \textbf{Sales} & \textbf{Codebase} & \textbf{Agg.} \\
\midrule
\textbf{RIKA} & Score & $\mathbf{0.387}\,\pm\,0.012$ & $\mathbf{0.952}\,\pm\,0.076$ & $-0.058\,\pm\,0.010$ & $\mathbf{0.434}\,\pm\,0.033$ & $0.628\,\pm\,0.040$ & $0.614\,\pm\,0.016$ & $\mathbf{0.130}\,\pm\,0.022$ \\
 & Gain & $\mathbf{+0.298}\,\pm\,0.012$ & $\mathbf{-0.688}\,\pm\,0.076$ & $+0.001\,\pm\,0.010$ & $\mathbf{+0.215}\,\pm\,0.033$ & $+0.189\,\pm\,0.040$ & $+0.207\,\pm\,0.016$ & $\mathbf{+0.201}\,\pm\,0.022$ \\
\addlinespace[2pt]
ICL & Score & $0.376\,\pm\,0.008$ & $0.790\,\pm\,0.069$ & $\mathbf{-0.034}\,\pm\,0.004$ & $0.367\,\pm\,0.025$ & $\mathbf{0.707}\,\pm\,0.035$ & $0.390\,\pm\,0.023$ & $0.084\,\pm\,0.015$ \\
 & Gain & $+0.287\,\pm\,0.008$ & $-0.850\,\pm\,0.069$ & $\mathbf{+0.025}\,\pm\,0.004$ & $+0.148\,\pm\,0.025$ & $\mathbf{+0.268}\,\pm\,0.035$ & $-0.017\,\pm\,0.023$ & $+0.160\,\pm\,0.015$ \\
\addlinespace[2pt]
Mem0 & Score & $0.355\,\pm\,0.015$ & $0.863\,\pm\,0.123$ & $-0.038\,\pm\,0.009$ & $0.334\,\pm\,0.050$ & $0.544\,\pm\,0.038$ & $0.583\,\pm\,0.010$ & $0.084\,\pm\,0.022$ \\
 & Gain & $+0.266\,\pm\,0.015$ & $-0.777\,\pm\,0.123$ & $+0.021\,\pm\,0.009$ & $+0.115\,\pm\,0.050$ & $+0.105\,\pm\,0.038$ & $+0.176\,\pm\,0.010$ & $+0.153\,\pm\,0.022$ \\
\addlinespace[2pt]
ACE & Score & $0.239\,\pm\,0.026$ & $0.452\,\pm\,0.103$ & $-0.074\,\pm\,0.012$ & $0.378\,\pm\,0.041$ & $0.591\,\pm\,0.067$ & $0.517\,\pm\,0.025$ & $0.020\,\pm\,0.025$ \\
 & Gain & $+0.150\,\pm\,0.026$ & $-1.188\,\pm\,0.103$ & $-0.015\,\pm\,0.012$ & $+0.159\,\pm\,0.041$ & $+0.152\,\pm\,0.067$ & $+0.110\,\pm\,0.025$ & $+0.101\,\pm\,0.025$ \\
\addlinespace[2pt]
Trace2Skill & Score & $0.098\,\pm\,0.027$ & $0.539\,\pm\,0.129$ & $-0.095\,\pm\,0.008$ & $0.395\,\pm\,0.052$ & $0.538\,\pm\,0.052$ & $\mathbf{0.622}\,\pm\,0.023$ & $-0.005\,\pm\,0.026$ \\
 & Gain & $+0.009\,\pm\,0.027$ & $-1.101\,\pm\,0.129$ & $-0.036\,\pm\,0.008$ & $+0.176\,\pm\,0.052$ & $+0.099\,\pm\,0.052$ & $\mathbf{+0.215}\,\pm\,0.023$ & $+0.079\,\pm\,0.026$ \\
\addlinespace[2pt]
SkillOpt & Score & $0.103\,\pm\,0.025$ & $0.588\,\pm\,0.114$ & $-0.087\,\pm\,0.010$ & $0.385\,\pm\,0.065$ & $0.559\,\pm\,0.048$ & $0.605\,\pm\,0.022$ & $0.004\,\pm\,0.029$ \\
 & Gain & $+0.014\,\pm\,0.025$ & $-1.052\,\pm\,0.114$ & $-0.028\,\pm\,0.010$ & $+0.166\,\pm\,0.065$ & $+0.120\,\pm\,0.048$ & $+0.198\,\pm\,0.022$ & $+0.086\,\pm\,0.029$ \\
\bottomrule
\end{tabular}%
}
\caption{\textbf{Gemini-3-Flash results with 95\% confidence intervals.} Means match Table~\ref{tab:main-results}. Task-level Gain intervals condition on the fixed shared control.}
\label{tab:main-ci-gemini}
\end{table*}

\section{Additional Ablations and Sensitivity Analyses}
\label{sec:appendix-additional-analysis}

All experiments in this section use DeepSeek-V4-Flash on \textsc{DB}, \textsc{Cohort}, and \textsc{Codebase}, following the setup in Section~\ref{sec:experimental-setup}. Where reported, Gain values in Tables~\ref{tab:routing-ablation-full}--\ref{tab:threshold-sensitivity} use the same shared stateless controls as the main comparison (Table~\ref{tab:paired-stateless}): $c_{\textsc{DB}}=0.060$, $c_{\textsc{Cohort}}=-0.118$, and $c_{\textsc{Codebase}}=0.512$. For every variant, $\textsc{Gain}=\textsc{Score}-c_t$.

\subsection{Full Results for Main-Text Ablations}
\label{sec:appendix-full-ablation-results}

Tables~\ref{tab:routing-ablation-full} and~\ref{tab:knowledge-ablation-full} report the complete results corresponding to the score-only Tables~\ref{tab:routing-ablation} and~\ref{tab:knowledge-ablation} in the main text. \textsc{Score} is the raw stream-level task score, while \textsc{Gain} is the improvement relative to the shared stateless control in the same raw reward units.

\begin{table}[!htbp]
\centering
\small
\setlength{\tabcolsep}{3.2pt}
\resizebox{\columnwidth}{!}{
\begin{tabular}{l*{6}{c}}
\toprule
\multirow{2}{*}{\textbf{\textsc{Method}}}
& \multicolumn{2}{c}{\textbf{\textsc{DB}}}
& \multicolumn{2}{c}{\textbf{\textsc{Cohort}}}
& \multicolumn{2}{c}{\textbf{\textsc{Codebase}}} \\
\cmidrule(lr){2-3}
\cmidrule(lr){4-5}
\cmidrule(lr){6-7}
& Score & Gain & Score & Gain & Score & Gain \\
\midrule
\textbf{RIKA}
& \textbf{0.281} & \textbf{+0.221}
& \textbf{-0.042} & \textbf{+0.076}
& \textbf{0.575} & \textbf{+0.063} \\
Imm. Only
& 0.225 & +0.165
& -0.089 & +0.029
& 0.540 & +0.028 \\
Def. Only
& 0.265 & +0.205
& -0.070 & +0.048
& 0.557 & +0.045 \\
Random Router
& 0.249 & +0.189
& -0.084 & +0.034
& 0.549 & +0.037 \\
\bottomrule
\end{tabular}
}
\caption{\textbf{Full admission-routing ablation results.} The score-only version appears in Table~\ref{tab:routing-ablation}.}
\label{tab:routing-ablation-full}
\end{table}

\begin{table}[!htbp]
\centering
\small
\setlength{\tabcolsep}{3.0pt}
\resizebox{\columnwidth}{!}{
\begin{tabular}{l*{6}{c}}
\toprule
\multirow{2}{*}{\textbf{Method}}
& \multicolumn{2}{c}{\textbf{\textsc{DB}}}
& \multicolumn{2}{c}{\textbf{\textsc{Cohort}}}
& \multicolumn{2}{c}{\textbf{\textsc{Codebase}}} \\
\cmidrule(lr){2-3}
\cmidrule(lr){4-5}
\cmidrule(lr){6-7}
& Score & Gain & Score & Gain & Score & Gain \\
\midrule
\textbf{RIKA}
& \textbf{0.281} & \textbf{+0.221}
& \textbf{-0.042} & \textbf{+0.076}
& \textbf{0.575} & \textbf{+0.063} \\
Generic 1-type
& 0.213 & +0.153
& -0.087 & +0.031
& 0.541 & +0.029 \\
Flat 4-type
& 0.261 & +0.201
& -0.057 & +0.061
& 0.563 & +0.051 \\
$-$Factual
& 0.253 & +0.193
& -0.060 & +0.058
& 0.547 & +0.035 \\
$-$Conceptual
& 0.265 & +0.205
& -0.065 & +0.053
& 0.553 & +0.041 \\
$-$Procedural
& 0.270 & +0.210
& -0.073 & +0.045
& 0.549 & +0.037 \\
$-$Metacognitive
& 0.274 & +0.214
& -0.068 & +0.050
& 0.556 & +0.044 \\
\bottomrule
\end{tabular}
}
\caption{\textbf{Full ablations of knowledge types and hierarchical extraction.} The score-only version appears in Table~\ref{tab:knowledge-ablation}.}
\label{tab:knowledge-ablation-full}
\end{table}

\paragraph{Confidence intervals.}
Tables~\ref{tab:routing-ablation-ci} and~\ref{tab:knowledge-ablation-ci} supplement the corresponding main-text ablations with 95\% confidence intervals. Intervals follow the five-run Student-$t$ definition in Appendix~\ref{sec:appendix-runs}.

\begin{table}[!htbp]
\centering
\small
\setlength{\tabcolsep}{3pt}
\resizebox{\columnwidth}{!}{%
\begin{tabular}{lccc}
\toprule
\textbf{Method} & \textbf{\textsc{DB}} & \textbf{\textsc{Cohort}} & \textbf{\textsc{Codebase}} \\
\midrule
\textbf{RIKA} & $\mathbf{0.281}\,\pm\,0.016$ & $\mathbf{-0.042}\,\pm\,0.016$ & $\mathbf{0.575}\,\pm\,0.020$ \\
Imm. Only & $0.225\,\pm\,0.019$ & $-0.089\,\pm\,0.014$ & $0.540\,\pm\,0.016$ \\
Def. Only & $0.265\,\pm\,0.011$ & $-0.070\,\pm\,0.010$ & $0.557\,\pm\,0.021$ \\
Random Router & $0.249\,\pm\,0.024$ & $-0.084\,\pm\,0.019$ & $0.549\,\pm\,0.027$ \\
\bottomrule
\end{tabular}%
}
\caption{\textbf{Admission-routing ablations with 95\% confidence intervals.} Entries report mean Score $\pm$ 95\% confidence-interval half-width.}
\label{tab:routing-ablation-ci}
\end{table}

\begin{table}[!htbp]
\centering
\small
\setlength{\tabcolsep}{3pt}
\resizebox{\columnwidth}{!}{%
\begin{tabular}{lccc}
\toprule
\textbf{Method} & \textbf{\textsc{DB}} & \textbf{\textsc{Cohort}} & \textbf{\textsc{Codebase}} \\
\midrule
\textbf{RIKA} & $\mathbf{0.281}\,\pm\,0.016$ & $\mathbf{-0.042}\,\pm\,0.016$ & $\mathbf{0.575}\,\pm\,0.020$ \\
Generic 1-type & $0.213\,\pm\,0.021$ & $-0.087\,\pm\,0.019$ & $0.541\,\pm\,0.023$ \\
Flat 4-type & $0.261\,\pm\,0.010$ & $-0.057\,\pm\,0.013$ & $0.563\,\pm\,0.027$ \\
$-$Factual & $0.253\,\pm\,0.012$ & $-0.060\,\pm\,0.008$ & $0.547\,\pm\,0.019$ \\
$-$Conceptual & $0.265\,\pm\,0.012$ & $-0.065\,\pm\,0.007$ & $0.553\,\pm\,0.014$ \\
$-$Procedural & $0.270\,\pm\,0.017$ & $-0.073\,\pm\,0.012$ & $0.549\,\pm\,0.012$ \\
$-$Metacognitive & $0.274\,\pm\,0.007$ & $-0.068\,\pm\,0.006$ & $0.556\,\pm\,0.014$ \\
\bottomrule
\end{tabular}%
}
\caption{\textbf{Ablations of knowledge types and hierarchical extraction with 95\% confidence intervals.} Entries report mean Score $\pm$ 95\% confidence-interval half-width.}
\label{tab:knowledge-ablation-ci}
\end{table}

\subsection{Ablations on Structural Components}
\label{sec:analysis-structure}

We ablate three supporting mechanisms in RIKA: atomization, claim forgetting, and state-based rendering.

\begin{table}[!htbp]
\centering
\small
\setlength{\tabcolsep}{3pt}
\resizebox{\columnwidth}{!}{%
\begin{tabular}{lccc}
\toprule
\textbf{Method} & \textbf{\textsc{DB}} & \textbf{\textsc{Cohort}} & \textbf{\textsc{Codebase}} \\
\midrule
\textbf{RIKA} & $\mathbf{0.281}\,\pm\,0.016$ & $\mathbf{-0.042}\,\pm\,0.016$ & $\mathbf{0.575}\,\pm\,0.020$ \\
$-$Atomize & $0.267\,\pm\,0.013$ & $-0.059\,\pm\,0.010$ & $0.564\,\pm\,0.012$ \\
$-$Forget & $0.263\,\pm\,0.016$ & $-0.053\,\pm\,0.015$ & $0.568\,\pm\,0.020$ \\
Incremental write & $0.249\,\pm\,0.023$ & $-0.060\,\pm\,0.020$ & $0.567\,\pm\,0.027$ \\
\bottomrule
\end{tabular}%
}
\caption{\textbf{Ablations of atomization, forgetting, and state-based rendering.} Entries report mean Score $\pm$ 95\% confidence-interval half-width.}
\label{tab:structural-ablation}
\end{table}

As shown in Table~\ref{tab:structural-ablation}, removing any of these components consistently reduces performance. Without atomization, scores drop by $0.014$--$0.017$ on DB and COHORT, reflecting less precise relation matching over compound claims. Disabling forgetting also degrades performance, as stale unsupported pending claims accumulate in the state. Finally, replacing state-based rendering with incremental rewriting yields further drops, including $0.032$ on \textsc{DB}, showing the benefit of regenerating the knowledge file from the current accumulation state rather than repeatedly editing previous text.

\subsection{Sensitivity Analysis}
\label{sec:analysis-sensitivity}

We study RIKA's sensitivity to the batch size $n$, which controls how many completed trajectories are jointly processed during each state update, and the deferred admission threshold $k$, which controls the net evidence required for a deferred claim to be admitted after each batch. The immediate threshold remains fixed at $1$.

\paragraph{Batch size.}
Table~\ref{tab:batch-sensitivity} varies $n$ from 1 to 10.
\par
\begin{table}[!htbp]
\centering
\small
\setlength{\tabcolsep}{3pt}
\resizebox{\columnwidth}{!}{%
\begin{tabular}{lccc}
\toprule
\textbf{Batch Size} & \textbf{\textsc{DB}} & \textbf{\textsc{Cohort}} & \textbf{\textsc{Codebase}} \\
\midrule
$n=1$ & $0.257\,\pm\,0.018$ & $-0.058\,\pm\,0.009$ & $0.567\,\pm\,0.017$ \\
$n=3$ & $\mathbf{0.285}\,\pm\,0.021$ & $\mathbf{-0.039}\,\pm\,0.018$ & $0.574\,\pm\,0.014$ \\
$n=5$ & $0.281\,\pm\,0.016$ & $-0.042\,\pm\,0.016$ & $\mathbf{0.575}\,\pm\,0.020$ \\
$n=10$ & $0.277\,\pm\,0.016$ & $-0.046\,\pm\,0.013$ & $0.573\,\pm\,0.016$ \\
\bottomrule
\end{tabular}%
}
\caption{\textbf{Sensitivity to batch size $n$.} Entries report mean Score $\pm$ 95\% confidence-interval half-width.}
\label{tab:batch-sensitivity}
\end{table}

\par
Processing one trajectory at a time consistently underperforms batched extraction, with drops of $0.024$ and $0.016$ relative to $n=5$ on \textsc{DB} and \textsc{Cohort}, respectively. A single trajectory provides less context for distinguishing reusable regularities from task-specific observations or accidental correlations. Performance is otherwise stable for $n=3$--$10$, suggesting that cross-trajectory context is useful but large batches are unnecessary.

\paragraph{Admission threshold.}
Table~\ref{tab:threshold-sensitivity} varies $k$ from 2 to 5.
\par
\begin{table}[!htbp]
\centering
\small
\setlength{\tabcolsep}{3pt}
\resizebox{\columnwidth}{!}{%
\begin{tabular}{lccc}
\toprule
\textbf{Threshold} & \textbf{\textsc{DB}} & \textbf{\textsc{Cohort}} & \textbf{\textsc{Codebase}} \\
\midrule
$k=2$ & $0.267\,\pm\,0.008$ & $-0.046\,\pm\,0.015$ & $0.574\,\pm\,0.015$ \\
$k=3$ & $\mathbf{0.281}\,\pm\,0.016$ & $-0.042\,\pm\,0.016$ & $0.575\,\pm\,0.020$ \\
$k=4$ & $0.279\,\pm\,0.013$ & $\mathbf{-0.038}\,\pm\,0.011$ & $\mathbf{0.576}\,\pm\,0.029$ \\
$k=5$ & $0.278\,\pm\,0.021$ & $-0.040\,\pm\,0.007$ & $0.575\,\pm\,0.023$ \\
\bottomrule
\end{tabular}%
}
\caption{\textbf{Sensitivity to admission threshold $k$.} Entries report mean Score $\pm$ 95\% confidence-interval half-width.}
\label{tab:threshold-sensitivity}
\end{table}

\par
Performance is generally stable across the evaluated thresholds. These results support the use of accumulated evidence from multiple trajectories for admission.

\section{Computational Cost}
\label{sec:appendix-cost}

Table~\ref{tab:main-results-cost} reports the costs in USD corresponding to the main results in Table~\ref{tab:main-results}.

\begin{table}[!htbp]
\centering
\small
\setlength{\tabcolsep}{2pt}
\begin{tabular}{lrrrr}
\toprule
\textbf{System} & \textbf{GPT-5.4} & \shortstack{\textbf{DeepSeek}\\\textbf{V4 Flash}} & \shortstack{\textbf{Gemini}\\\textbf{3 Flash}} & \textbf{MiniMax-M3} \\
\midrule
ICL & \$16.32 & \$0.75 & \$3.22 & \$1.60 \\
Mem0 & \$16.43 & \$0.64 & \$3.35 & \$1.57 \\
ACE & \$62.35 & \$2.33 & \$11.98 & \$6.44 \\
Trace2Skill & \$23.74 & \$1.03 & \$5.21 & \$2.35 \\
SkillOpt & \$46.33 & \$2.15 & \$8.33 & \$4.57 \\
\textbf{RIKA} & \$25.89 & \$1.21 & \$5.11 & \$2.53 \\
\bottomrule
\end{tabular}
\caption{\textbf{Costs corresponding to the main results.} Costs are reported in USD for the systems and model backbones evaluated in Table~\ref{tab:main-results}.}
\label{tab:main-results-cost}
\end{table}

\end{document}